\documentclass[12pt]{article}
\usepackage{amsmath,amsfonts,amsthm,amssymb}
\usepackage{graphicx,color,lscape,hyperref}
\usepackage[numbers]{natbib}
\theoremstyle{definition}
\newtheorem{definition}{Definition}
\newtheorem{proposition}{Proposition}
\newtheorem{corollary}{Corollary}
\newtheorem{lemma}{Lemma}

\begin{document}
\title{Hybrid Fuzzy-Crisp Clustering Algorithm: Theory and Experiments}
\author{Akira R. Kinjo$^1$ and Daphne Teck Ching Lai$^2$\\
  $^1$Department of Mathematics and $^2$School of Digital Science,\\
  University Brunei Darussalam}
\maketitle

\begin{abstract}
  With the membership function being strictly positive, the conventional fuzzy c-means clustering method sometimes causes imbalanced influence when clusters of vastly different sizes exist. That is, an outstandingly large cluster drags to its center all the other clusters, however far they are separated. To solve this problem, we propose a hybrid fuzzy-crisp clustering algorithm based on a target function combining linear and quadratic terms of the membership function. In this algorithm, the membership of a data point to a cluster is automatically set to exactly zero if the data point is ``sufficiently'' far from the cluster center. In this paper, we present a new algorithm for hybrid fuzzy-crisp clustering along with its geometric interpretation. The algorithm is tested on twenty simulated data generated and five real-world datasets from the UCI repository and compared with conventional fuzzy and crisp clustering methods. The proposed algorithm is demonstrated to outperform the conventional methods on imbalanced datasets and can be competitive on more balanced datasets.
\end{abstract}

\section{Introduction}
Fuzzy clustering is a clustering method in which each data point may belong to multiple clusters \citep{Ruspini2019fuzzy}. The membership of a data point to clusters is expressed in terms of a membership function.
In the conventional fuzzy c-means clustering (FCM) method \citep{Dunn1973FuzzyClustering,Bezdek1984fcm}, the membership function is strictly positive. This sometimes causes a problem when there are clusters of widely different sizes. For example, consider a set of samples drawn from the following Gaussian mixture (Fig. \ref{fig:twoclust}):
\begin{figure}[h]
  \begin{center}
    \includegraphics[width=0.6\textwidth]{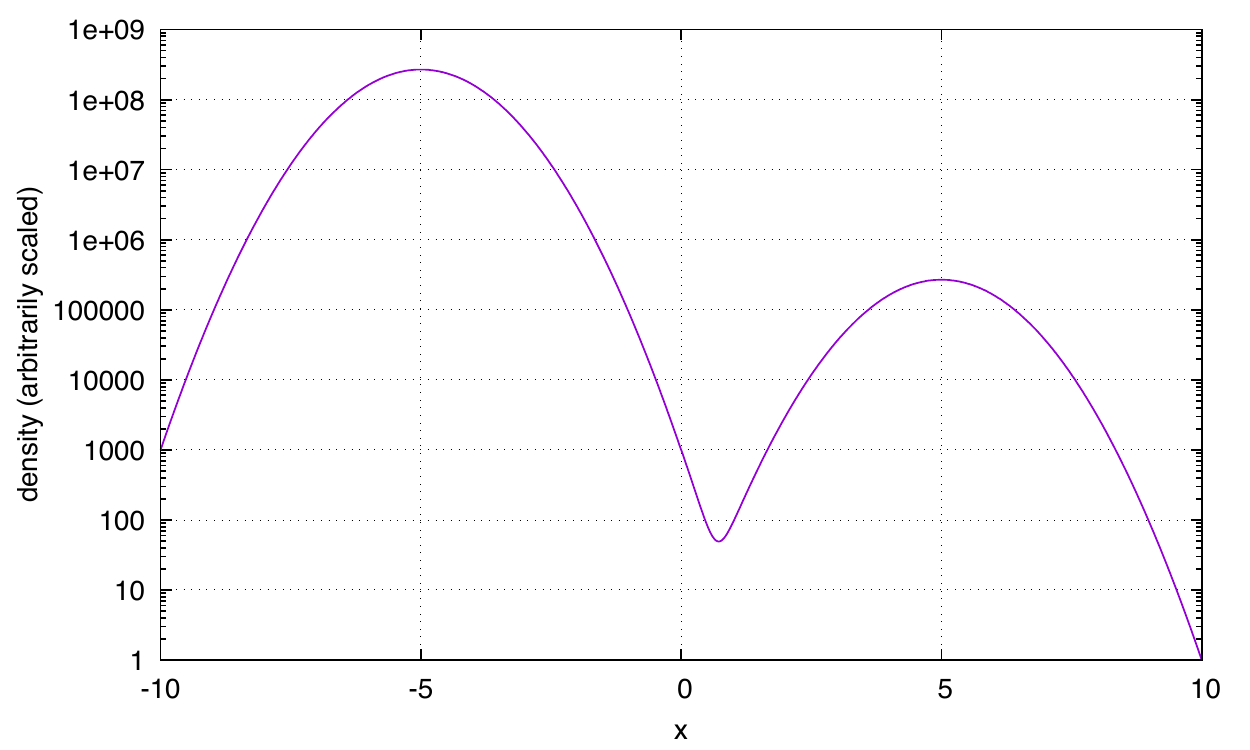}
  \end{center}
  \caption{\label{fig:twoclust}An example of two clusters of very different sizes (c.f. Eqs. \ref{eq:gmix1} and \ref{eq:gauss}).}
  
\end{figure}
\begin{equation}
  \rho(x) \propto 10^3\phi(x; -5, 1) + \phi(x; +5, 1) \label{eq:gmix1}
\end{equation}
where the Gaussian density is given as
\begin{equation}
  \phi(x; \mu, \sigma) = \frac{1}{\sqrt{2\pi\sigma^2}}e^{-\frac{(x-\mu)^2}{2\sigma^2}}. \label{eq:gauss}
\end{equation}
There are two clearly separated clusters (i.e., $c=2$), and one (Cluster 1) of them is 1,000 times more populated than the other (Cluster 2).
Now, let's apply the conventional fuzzy c-means clustering to this data set. That is, we try to minimize the following objective function:
\begin{equation}
  J(\{u_i(x)\}, \{v_i\}) = \sum_{i=1}^{c}\int_{V}dx\rho(x)[u_i(x)]^m\|x - v_i\|^2
\end{equation}
where $V$ is the range of data point $x$ (in the present case, $V = (-\infty, +\infty)$), $m>1$ is a constant, $v_i$ is the coordinate of the $i$-th cluster center, and $u_i(x)$ is the membership of the point $x$ to the $i$-th cluster. This objective function can be optimized by iteratively applying the following mutually recursive formulae:
\begin{eqnarray}
  u_i(x) &\gets& \left(\sum_{j=1}^{c}\left(\frac{\|x - v_i\|^2}{\|x - v_j\|^2}\right)^{\frac{1}{m-1}}\right)^{-1},\\
  v_i &\gets& \frac{\int_{V}dx\rho(x)[u_i(x)]^mx}{\int_{V}dx\rho(x)[u_i(x)]^m}.
\end{eqnarray}
It is noted that the fuzzy c-means clustering reduces to the crisp c-means (k-means) clustering in the limit $m \to 1+0$.
With the data set following $\rho(x)$ in Eq. (\ref{eq:gmix1}) and the initial cluster centers given as $v_1 = -5$ and $v_2 = +5$, the conventional fuzzy c-means clustering yields the result shown in Fig. \ref{fig:example1}A and B.
We can see that the center of Cluster 2 is attracted towards the center of Cluster 1 on each update, and at convergence, we have $v_1 \approx -5.80$ and $v_2 \approx -4.17$. It is clear that the points with large membership values to Cluster 2 are buried within Cluster 1, and the membership at around $x = +5$ (the center of the smaller Gaussian) is very fuzzy with $u_1(5) \approx 0.42$ and $u_2(5) \approx 0.58$.
\begin{figure}
  \begin{center}
    \includegraphics[width=1.0\textwidth]{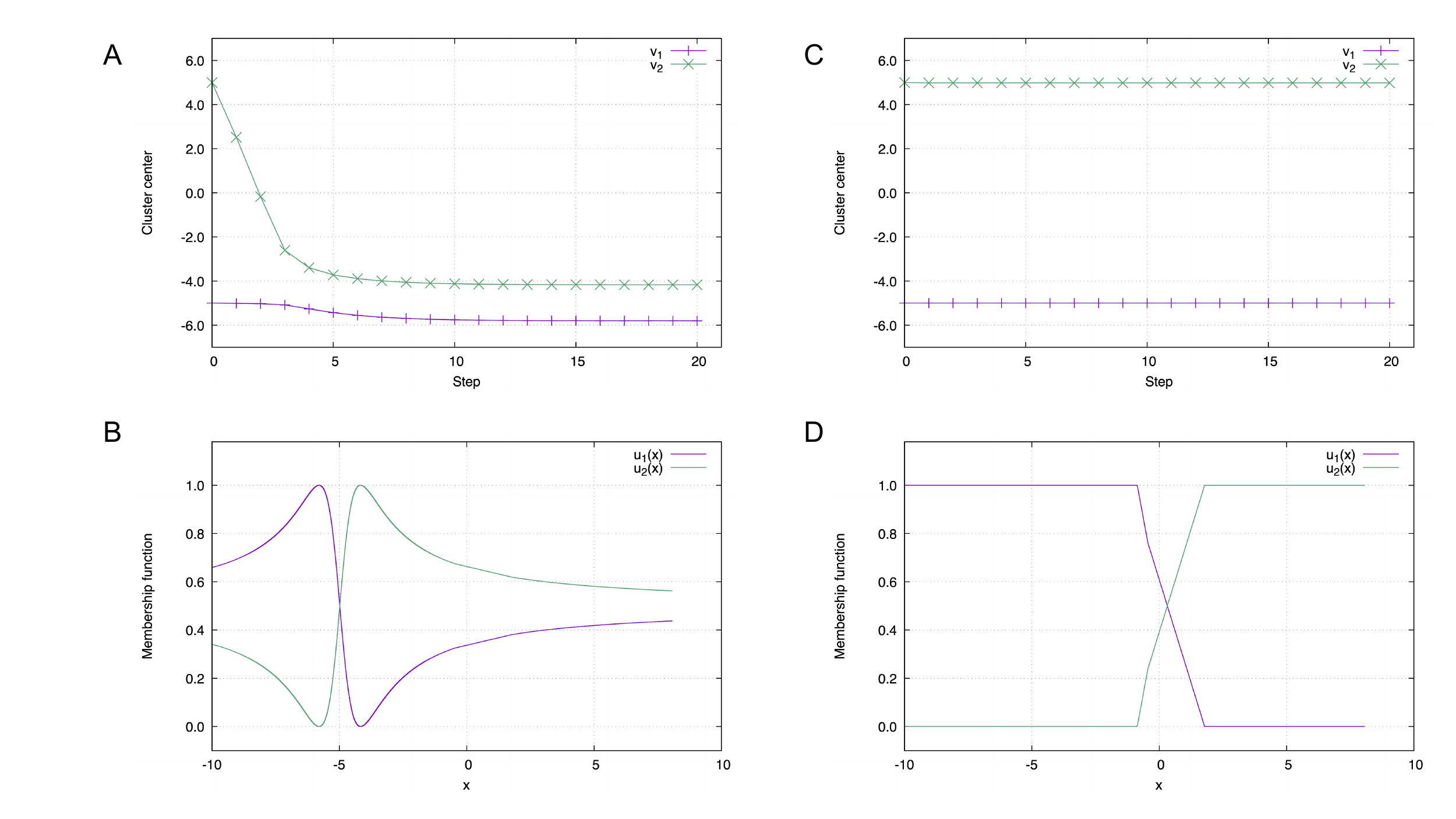}
  \end{center}
  \caption{\label{fig:example1}Applications of fuzzy clustering to the data generated by the Gaussian mixture (Eq. \ref{eq:gmix1}, Fig. \ref{fig:twoclust}).
    A and B are based on the conventional FCM method. C and D are based on the Hybrid Fuzzy-Crisp (HFC) method.
    A: Trajectory of cluster centers according to FCM.
    B: Membership functions obtained by FCM with initial cluster centers being the means of the mixed Gaussians.
    C: Trajectory of cluster centers according to HFC.
    D: The membership functions obtained by HFC with initial cluster centers being the means of the mixed Gaussians.}
\end{figure}

As we have already mentioned, the source of the above problem is that the membership functions assume strictly positive values. Thus, when one cluster is outstandingly large, it will drag the centers of the other clusters however far they are separated.

One way to alleviate this problem is to set the membership of a data point to a cluster to exactly zero if the data point is ``sufficiently'' far from the cluster center. \citet{KinjoMaster} has shown that this is indeed possible by using an objective function that combines linear and quadratic terms of the membership functions. However, his proposed algorithm was rather primitive and extremely inefficient (exponential in the number of clusters). Later, \citet{klawonn2003fuzzy} independently proposed essentially the same objective function, but with a more efficient, practical algorithm.
In the simplest version of this hybrid fuzzy-crisp clustering method (HFC), the objective function to be minimized is given as
\begin{equation}
  J(\{u_i(x)\}, \{v_i\}) = \sum_{i=1}^{c}\int_{V}dx\rho(x)\left(\frac{u_i^2(x)}{2} + u_{i}(x)\right)\|x - v_i\|^2
\end{equation}
where $u_i^2(x) = [u_i(x)]^2$.
We will provide a more general objective function below.
If we apply HFC to the distribution given by Eq. (\ref{eq:gmix1}), we obtain the result shown in Fig. \ref{fig:example1}C, D. The center of the smaller cluster is no longer attracted to the larger cluster. In fact, the centers converge at $v_1 \approx -5.00$ and $v_2 \approx +4.98$ (Fig. \ref{fig:example1}C). Furthermore, the membership functions can take values exactly equal to 1 or 0 near cluster centers (Fig. \ref{fig:example1}D).
As we can observe in this example, HFC combines characteristics of fuzzy and crisp (hard) clustering methods thereby solving some difficulties in the conventional clustering methods.

In this paper, we present a new algorithm for the fuzzy-crisp clustering based on a geometric interpretation of the objective function minimization problem.

\section{Theory}
\subsection{Hybrid Fuzzy-Crisp c-means clustering (HFC)}
Suppose we have the data distribution $\rho(x)$ with $x\in \mathbb{R}^d, x \in V$ where $V$ is the range of the data points. When the data consist of $n$ discrete points, $x_1, x_2, \cdots, x_n$, the distribution can be written as $\rho(x) = \frac{1}{n}\sum_{k=1}^{n}\delta(x - x_k)$ where $\delta$ is the Dirac's delta function.
We assume that there are $c$ clusters whose centers are $v_1, v_2, \cdots, v_c \in \mathbb{R}^d$. The membership of the data point $x$ to cluster $i$ is represented by the membership functions $u_i(x)$. We define the distance function between data point $x$ and cluster center $v_i$ by the Euclidean norm $d_i(x) = \|x - v_i\|$. In the following, we denote the square of the function value $f(x)$ by $f^2(x)$ (e.g., $u_i^2(x)$ indicates $[u_i(x)]^2$). Our task is to find the optimal cluster centers $\{v_i\}$ and membership functions $\{u_i(x)\}$ that minimize the following objective function
\begin{equation}
  J(\{u_i(x)\},\{v_i\}) = \sum_{i=1}^{c}\int_V dx \rho(x)\left(\frac{u_{i}^2(x)}{2\alpha_i} + \frac{u_{i}(x)}{\beta}\right)d_{i}^2(x)\label{eq:JFCC}
\end{equation}
with the constraints
\begin{equation}
  u_{i}(x) \geq 0, i = 1, \cdots, c \label{eq:uconst1}
\end{equation}
and
\begin{equation}
  \sum_{i=1}^{c}u_{i}(x) = 1 \label{eq:uconst2}
\end{equation}
for all $x \in V$.
$\alpha_i$'s and $\beta$ are positive parameters, and we impose the constraint
\begin{eqnarray}
  \sum_{i=1}^{c}\alpha_i  &=& 1.\label{eq:aconst}
\end{eqnarray}
We assume that $\beta$ is given and fixed. $\alpha_i$'s may be either fixed or optimized. We have found that the weight of the linear term ($\beta$) cannot be cluster-dependent (such as $\{\beta_i\}$) as it may cause the membership functions to explode.

Given the membership functions $\{u_{i}(x)\}$, the optimal cluster centers $\{v_{i}\}$ are obtained by solving
\begin{equation}
  \frac{\partial J}{\partial v_i} = 0,
\end{equation}
which gives
\begin{equation}
  v_i = \frac{\int_Vdx\rho(x)({u_{i}^2(x)}/{2\alpha_i} + u_{i}(x)/\beta)x}{\int_Vdx\rho(x)({u_{i}^2(x)}/{2\alpha_i} + u_{i}(x)/\beta)}.
\end{equation}

Optimizing the objective function with respect to $\alpha_i$'s with the constraint (Eq. \ref{eq:aconst}), with all the other variables being fixed, gives
\begin{equation}
  \alpha_i = \frac{\sqrt{\int_{V}dx\rho(x)u_i^2(x)d_i^2(x)}}{\sum_{j=1}^{c}\sqrt{\int_{V}dx\rho(x)u_j^2(x)d_j^2(x)}}.
\end{equation}
Given the cluster centers $\{v_i\}$, the objective function $J$ is minimized with constraints Eqs. (\ref{eq:uconst1}) and (\ref{eq:uconst2}). This minimization problem can be solved with a geometric method (see the next section).
First, we need the following result:
\begin{proposition}\label{prop:H}
For each $x\in V$, there exists a unique subset of clusters $H \subset C = \{1, 2, \cdots, c\}$ such that
  \begin{eqnarray}
    {d_{i}^2(x)} &<& d_{H}^2(x)~~ \text{if $i \in H$}, \label{eq:ineqdH}\\
    {d_{i}^2(x)} &\geq& d_{H}^2(x)~~ \text{if $i \notin H$}\label{eq:ineqdHc}
  \end{eqnarray}
where
\begin{equation}
  d_{H}^2(x) = \frac{\beta + \sum_{j\in H}\alpha_j}{\sum_{j\in H}\alpha_j/d_{j}^2(x)}.\label{eq:dH}
\end{equation}
Using this subset $H$ of clusters, the optimal membership functions are given for each $x \in V$ as
\begin{equation}
  u_{i}(x) =\left\{
  \begin{array}{cc}
    \frac{\alpha_i}{\beta}\left(\frac{d_{H}^2(x)}{d_{i}^2(x)} - 1\right) & \text{if $i \in H$},\\
    0 & \text{if $i \notin H$}.
  \end{array}
  \right.\label{eq:memfunct}
\end{equation}
\end{proposition}
The proof is given in the subsection \ref{sec:geom} (\nameref{sec:geom}) below.

\subsection{Algorithm for finding the membership function of a given data point}
\label{sec:algo}
Here is an algorithm to compute the optimal membership function $\{u_{i}(x)\}_{i=1,\cdots, c}$ for a given data point $x \in V$.
\begin{enumerate}
\item Let $\nu = 0$ and $H^{(0)} = \{1, 2, \cdots, c\}$.
\item Calculate $d^2_{H^{(\nu)}}(x)$ by Eq. (\ref{eq:dH}).
\item If $d_{i}^{2}(x) < d^2_{H^{(\nu)}}(x)$ for all $i \in H^{(\nu)}$, then
  set
\begin{equation}
  u_{i}(x) =\left\{
  \begin{array}{cc}
    \frac{\alpha_i}{\beta}\left(\frac{d_{H^{(\nu)}}^2(x)}{d_{i}^2(x)} - 1 \right) & \text{if $i \in H^{(\nu)}$},\\
    0 & \text{if $i \notin H^{(\nu)}$}.
  \end{array}
  \right.
\end{equation}
and we are done. Otherwise, go to the next step.
\item Set
  \begin{equation}
    H^{(\nu+1)} = \{ i \in H^{(\nu)} \mid d_{i}^{2}(x) < d^2_{H^{(\nu)}}(x)\}.
  \end{equation}
\item Update $\nu := \nu + 1$ and go to Step 2.
\end{enumerate}
In Step 4, multiple clusters may be eliminated at once. This is in contrast to the algorithm by \citet{klawonn2003fuzzy} where only one cluster is eliminated at a time.

\subsection{Relationship to fuzzy c-means clustering}
We can show that HFC can recover FCM (with $m=2$) in the limit $\beta \to \infty$. In this limit, we have
\begin{equation}
  \lim_{\beta\to\infty}d_H^2 = \infty
\end{equation}
so that Eq. (\ref{eq:ineqdH}) is trivially satisfied for any subset $H \subset C$, in particular, for $H = C$. Furthermore, the membership function becomes
\begin{equation}
  u_i(x) = \lim_{\beta\to\infty}\frac{\alpha_i}{\beta}\left(\frac{d_H^2(x)}{d_i^2(x)} - 1\right) = \frac{\alpha_i/d_i^2}{\sum_{j\in H}\alpha_j/d_j^2(x)}
\end{equation}
which is the membership function of the FCM with $m=2$ (with the weighting factors $\{\alpha_i\}$).
%%%%%%%%%%%%%%%%%
\subsection{Analytical results with $c = 2$}\label{sec:c2}
\begin{figure}
  \begin{center}
    \includegraphics[width=10cm]{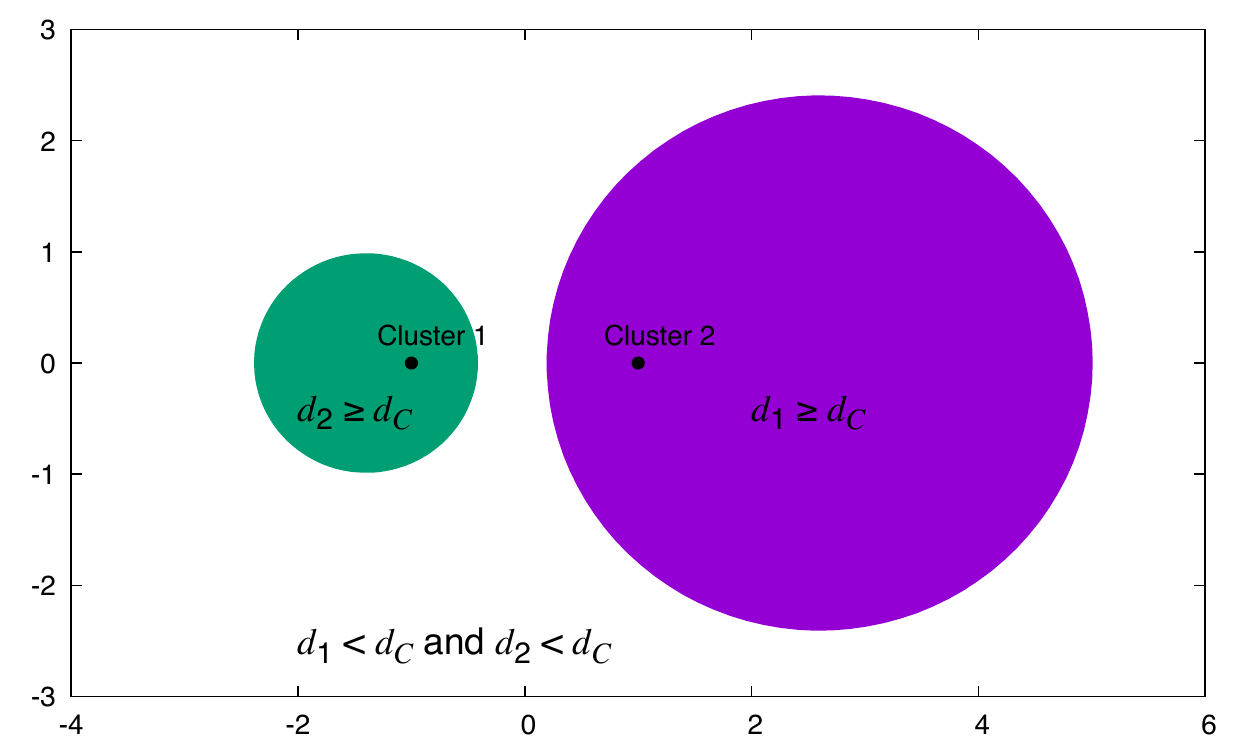}
  \end{center}
  \caption{\label{fig:c2}The case for $c=2$ with $v_1 = (-1, 0), v_2 = (1,0)$ and $\alpha_1 = 0.2, \alpha_2 = 0.8$. Two cluster centers are marked with black dots labeled ``Cluster 1'' and ``Cluster 2''. The green circle indicates the area where $d_2 \geq d_C$ so that $u_1 = 1, u_2 = 0$; the magenta circle indicates the area where $d_1 \geq d_C$ so that $u_1 = 0, u_2 = 1$; the area outside the two circles indicates that with $u_1 > 0, u_2 > 0$.}
\end{figure}

We have already given an example with $c=2$ (two clusters) in Fig. \ref{fig:example1} (C and D). Here, we illustrate in detail the case with $c=2$ with fixed cluster centers.

Suppose that there are two cluster centers in $\mathbb{R}^2$, one at $v_1 = (-x_0,0)$ and the other at $v_2 = (x_0,0)$ with $x_0 > 0$. Given an arbitrary data point $(x,y)$, what are the membership functions? 

To simplify the notation, let us define the distances
\begin{eqnarray}
  d_1^2 &=& (x + x_0)^2 + y^2,\\
  d_2^2 &=& (x - x_0)^2 + y^2.
\end{eqnarray}
First, consider the case where $H = C = \{1, 2\}$. The threshold distance is
\begin{equation}
  d_C^2 = \frac{\beta + \alpha_1 + \alpha_2}{\alpha_1 / d_1^2 + \alpha_2 /d_2^2}.
\end{equation}

\paragraph{Case 1: $d_1^2 < d_C^2$ and $d_2^2 < d_C^2$.}
Then, the two inequalities
\begin{equation}
  \left\{
  \begin{array}{c}
    d_1^2 < d_C^2\\
    d_2^2 < d_C^2
  \end{array}\right.
\end{equation}
are equivalent to
\begin{eqnarray}
  \left[x - (1 + 2\alpha_2/\beta)x_0 \right]^2 + y^2 &>& 4(\alpha_2/\beta) (1 + \alpha_2/\beta)x_0^2 \label{eq:c1}\\
  \left[x + (1 + 2\alpha_1/\beta)x_0 \right]^2 + y^2 &>& 4(\alpha_1/\beta) (1 + \alpha_1/\beta)x_0^2 \label{eq:c2}.
\end{eqnarray}
Eq. (\ref{eq:c1}) corresponds to the exterior of the circle centered at $((1 + 2\alpha_2/\beta)x_0, 0)$ with radius $2x_0\sqrt{(\alpha_2/\beta)(1 + \alpha_2/\beta)}$, and Eq. (\ref{eq:c2}) to exterior of the circle centered at $(- (1 + 2\alpha_1/\beta)x_0, 0)$ with radius $2x_0\sqrt{(\alpha_1/\beta)(1+ \alpha_1/\beta)}$. Thus, when the data point $(x,y)$ is in the intersection of these regions, the membership functions $u_1$ and $u_2$ are both non-zero and are given by
\begin{eqnarray}
  u_1 &=& \frac{\alpha_1}{\beta}\left(d_C^2/d_1^2 - 1\right) = \alpha_1\left(\frac{\beta + \alpha_1 + \alpha_2}{\alpha_1 + \alpha_2d_1^2/d_2^2} - 1\right),\\
  u_2 &=& \frac{\alpha_2}{\beta}\left(d_C^2/d_2^2 - 1\right) = \alpha_2\left(\frac{\beta + \alpha_1 + \alpha_2}{\alpha_1d_2^2/d_1^2 + \alpha_2} - 1\right).
\end{eqnarray}
It is noted that we indeed have $u_1 > 0$, $u_2 > 0$ and, using $\alpha_1 + \alpha_2 = 1$,
\begin{equation}
  u_1 + u_2 = 1.
\end{equation}

\paragraph{Case 2: $d_1^2 < d_C^2$ and $d_2^2 \geq d_C^2$.}
In this case,  we set $H_1 = \{i \in C | d_i^2 < d_C^2\} = \{1 \}$ and
\begin{equation}
  d_{H_1}^2 = \frac{\beta + \alpha_1}{\alpha_1/d_1^2} = (1 + \beta/\alpha_1)d_1^2.\label{eq:h1d1}
\end{equation}
Let us see if we indeed have
\begin{eqnarray}
  d_1^2 &< & d_{H_1}^2,\\
  d_2^2 &\geq & d_{H_1}^2.
\end{eqnarray}
These inequalities are equivalent, respectively, to
\begin{eqnarray}
  d_1^2 &>& 0,\label{eq:c21}\\
  \left[x + (1 + 2\alpha_1/\beta)x_0 \right]^2 + y^2 &\leq& 4(\alpha_1/\beta) (1 + \alpha_1/\beta)x_0^2 \label{eq:c22}.
\end{eqnarray}
Eq. (\ref{eq:c21}) is trivially satisfied. Eq. (\ref{eq:c22}) is the complement of the region defined by Eq. (\ref{eq:c2}) which, in turn, follows from $d_2^2 < d_C^2$. Eq. (\ref{eq:c22}) follows from $d_2^2 \geq d_C^2$, and hence is satisfied. The membership function is given by
\begin{eqnarray}
  u_1 &=& \alpha_1(d_{H_1}^2/d_1^2 - 1) = 1,\\
  u_2 &=& 0.
\end{eqnarray}

\paragraph{Case 3: $d_1^2 \geq d_C^2$ and $d_2^2 < d_C^2$.}
By symmetry, we set $H_2 = \{2 \}$ and define
\begin{equation}
  d_{H_2}^2 = \frac{\beta + \alpha_2}{\alpha_2/d_2^2} = (1 + \beta/\alpha_2)d_2^2.
\end{equation}
Thus, we have
\begin{eqnarray}
  d_1^2 & \geq & d_{H_2}^2,\label{eq:c31}\\
  d_2^2 &< & d_{H_2}^2\label{eq:c32}
\end{eqnarray}
and
\begin{eqnarray}
  u_1 &=& 0,\\
  u_2 &=& 1.
\end{eqnarray}

\paragraph{Case 4: $d_1^2 \geq d_C^2$ and $d_2^2 \geq d_C^2$.}
This case cannot happen.
The distance between the centers of the circles defined by $d_1^2 = d_C^2$ and $d_2^2 = d_C^2$ is
\begin{equation}
(1 + 2\alpha_2/\beta)x_0 + (1 + 2\alpha_1/\beta)x_0 = 2x_0(1 + \alpha_1/\beta + \alpha_2/\beta) = 2x_0(1 + 1/\beta).
\end{equation}
For any positive real numbers $a$ and $b$, we have the inequality
\begin{equation}
  \sqrt{ab} \leq \frac{a + b}{2}.
\end{equation}
Using this, the sum of the radii of the two circles, divided by $2x_0$, is
\begin{equation}
  \sqrt{(\alpha_2/\beta)(1 + \alpha_2/\beta)} + \sqrt{(\alpha_1/\beta)(1 + \alpha_1/\beta)} \leq 1 + 1/\beta.
\end{equation}
Thus the sum of the radii is smaller than the distance between the centers, and hence the two circles cannot intersect.

An example is shown in Fig. \ref{fig:c2} where we set $\alpha_1 = 0.2$ and $\alpha_2 = 0.8$. The colored areas indicate the regions where one of the membership functions is exactly 0.

%%%%%%%%%%%%%%%%%
%%%%%%%%%%%%%%%%%

\subsection{Geometric interpretation}\label{sec:geom}
Here, we show that the problem of finding the optimal membership function $u_{i}(x)$ for a single data point $x$ can be interpreted as finding the point in a $(c-1)$-simplex in the $c$-dimensional space that is closest to the origin. By solving the latter problem, we prove Proposition \ref{prop:H}. We will omit the argument ``$x$'' henceforth. Thus the objective function is
\begin{equation}
  J(\{u_i\}) = \sum_{i=1}^{c}\left(\frac{u_{i}^2}{2\alpha_i} + \frac{u_{i}}{\beta}\right)d_{i}^2
\end{equation}
with constraints
\begin{equation}
  u_{i} \geq 0
\end{equation}
and
\begin{equation}
  \sum_{i=1}^{c}u_{i} = 1.
\end{equation}
Now define
\begin{eqnarray}
  a_i &=& \frac{d_{i}}{\sqrt{2\alpha_i}},\\
  b_i &=& \sqrt{\frac{\alpha_i}{2}}\frac{d_{i}}{\beta},\\
  \xi_i &=& a_iu_i + b_i.\label{eq:xi}
\end{eqnarray}
Then the objective function is represented as
\begin{equation}
  J(\{\xi_i\}) = \sum_{i=1}^{c}(\xi_i^2 - b_i^2)\label{eq:objxi}
\end{equation}
with the constraints
\begin{equation}
  \xi_i \geq b_i, i = 1, 2, \cdots, c, \label{eq:xirange}
\end{equation}
and
\begin{equation}
  \sum_{i=1}^{c}\frac{\xi_i - b_i}{a_i} = 1.\label{eq:plane}
\end{equation}
Eq. (\ref{eq:plane}) defines a hyperplane in $\mathbb{R}^c$ that contains points $P_1, P_2, \cdots, P_c$ where the $j$-th coordinate of point $P_i$ is
\begin{equation}
(P_i)_j = a_j\delta_{i,j} + b_j.
\end{equation}
Here, $\delta_{i,j}$ is Kronecker's delta: $\delta_{i,j} = 1$ if $i = j$, or $\delta_{i,j} = 0$, otherwise.
Together with the restraints Eq. (\ref{eq:xirange}), it defines the $(c-1)$-simplex consisting of vertices $P_1, P_2, \cdots, P_c$. Since $b_i$'s are constants, minimizing the objective function $J(\{\xi_i\})$ is equivalent to finding the point in the $(c-1)$-simplex that is closest to the origin.

Let $C = \{1, 2, \cdots, c\}$ be the set of indices of vertices. For a subset $S \subset C$, let $\pi(S)$ denote the hyperplane defined by the points $\{P_i\}_{i\in S}$. Similarly, let $\mathcal{S}(S)$ denote the $(|S|-1)$-simplex with vertices $(P_i)_{i\in S}$. $\mathcal{S}(S)$ is a facet of the simplex $\mathcal{S}(C)$. It is clear that the closest point from the origin to the simplex $\mathcal{S}(C)$ resides in one of the facets of the simplex (including $\mathcal{S}(C)$). 
\begin{definition}[Optimal subset]
  We say that a subset $H \subset C$ is \emph{optimal} if the closest point from the origin to the simplex $\mathcal{S}(C)$ exists in the interior of the facet $\mathcal{S}(H)$. 
\end{definition}
Note, in particular, that $H$ is \emph{not} optimal if the closest point from the origin exists on the boundary of $\mathcal{S}(H)$.

Now let $G$ be the foot of the perpendicular to $\pi(C)$ from the origin $O$. Then, the coordinates of $G = (\xi_i^G)$ are given as
\begin{equation}
  \xi_i^G = \frac{1 + \sum_{j\in C}\frac{b_j}{a_j}}
     {a_i\sum_{j\in C}\frac{1}{a_j^2}}.\label{eq:G}
\end{equation}
If $\xi_i^G > b_i$ for all $i\in C$, then this $G$ is in the \emph{interior} of $\mathcal{S}(C)$ and it is the closest point from the origin. If $\xi_i^G \leq b_i$ for some $i \in C$ and $\xi_i^G > 0$ for all the other $i\in C$, then the point $G$ is outside or on one of the facets of the simplex $\mathcal{S}(C)$, and the closest point $G^{*}$ must exist on one (and only one) of the facets of $\mathcal{S}(C)$. Let the the set of vertices of that facet be $H \subset C$. A facet of a simplex is again a simplex (in a lower dimension). Thus, the point $G^*$ is an interior point of simplex $\mathcal{S}(H)$.
We define the following quantities for convenience:
\begin{eqnarray}
  B &=& (b_1, b_2, \cdots, b_c),\\
  \mathbf{p}_i &=& \overrightarrow{BP}_i = (a_i\delta_{i,j})_{j=1,\cdots,c},\\
  A_S^2 &=& \left(\sum_{i\in S}\frac{1}{a_i^2}\right)^{-1} \text{  for subset $S \subset C$},\\
  D_S &=& \sum_{i\in S}\frac{b_i}{a_i} \text{  for subset $S \subset C$},\\
  \mathbf{n} &=& \sum_{i=1}^{c}\frac{A_C^2}{a_i^2}\mathbf{p}_i,\label{eq:ndef}\\
  \mathbf{q}_i &=& \mathbf{n} - \mathbf{p}_i\label{eq:qdef}.
\end{eqnarray}
For any $i, j,k \in C$, we have
\begin{equation}
  \overrightarrow{P_jP_k}\cdot \mathbf{q}_i = a_i^2(\delta_{ij} - \delta_{ik}).\label{eq:qperp}
\end{equation}
Thus, for any subset $H\subset C$, the subspaces spanned by $\{\mathbf{p}_i\}_{i\in H}$ and by $\{\mathbf{q}_i\}_{i\in H^c}$ are orthogonal to each other ($H^c = C\setminus H$ is the complement of $H$).

\begin{lemma}
  Let $H$ be the optimal subset of $C$, and $G$ be the foot of the perpendicular to the hyperplane $\pi(C)$ from the origin (Eq. \ref{eq:G}). Then, we have
  \begin{equation}
    \overrightarrow{BG} = \sum_{i\in H}s_i\mathbf{p}_i + \sum_{i\in H^c}s_i'\mathbf{q}_i\label{eq:Hrep}
  \end{equation}
  where $\sum_{i\in H}s_i = 1$, $s_i > 0$ for all $i\in H$, $s_i' \geq 0$ for all $i\in H^c$.
\end{lemma}
\begin{proof}
  There are three cases.
  \paragraph{Case 1: $G$ is in the interior of $\mathcal{S}(C)$}
  In this case, $H = C$ and $H^c = \emptyset$. Eq. (\ref{eq:Hrep}) clearly holds
  with $\sum_{i\in C}s_i = 1$, and $s_i > 0$ for all $i \in C$ (the second sum over $i\in H^c (= \emptyset)$ is absent).
  \paragraph{Case 2: $G$ is in the interior of the facet $\mathcal{S}(H)$ and $H\subsetneq C$}
  In this case, Eq. (\ref{eq:Hrep}) holds with $\sum_{i\in H}s_i= 1$,  $s_i > 0$ for all $i \in H$, and $s_i' = 0$ for all $i \in H^c$.
  \paragraph{Case 3: $G$ is in the exterior of $\mathcal{S}(C)$}
  Let $G^*$ be the closest point in $\mathcal{S}(C)$ from the origin. Then, $G^{*}$ is in the interior of some facet $\mathcal{S}(H)$ where $H \subsetneq C$.
  We have
  \begin{equation*}
    \overrightarrow{BG} = \overrightarrow{BG^{*}} + \overrightarrow{G^{*}G}.
  \end{equation*}
  Since $G^{*}$ is in the interior of $\mathcal{S}(H)$, we have
\begin{equation}
\overrightarrow{BG^*} = \sum_{i\in H}s_i \mathbf{p}_i\label{eq:bgstar}
\end{equation}
where $\sum_{i\in H}s_i = 1$ and $s_i > 0$ for all $i\in H$.
Furthermore, the vector $\overrightarrow{G^{*}G}$ is perpendicular to the facet $\mathcal{S}(H)$ (i.e., $G^{*}$ is the point on the hyperplane $\pi(H)$ that is closest to $G$). Noting that the subspace spanned by  $\{\mathbf{q}_i\}_{i \in H^{c}}$ is orthogonal to the subspace $\pi(H)$ (Eq. \ref{eq:qperp}), the vector $\overrightarrow{G^{*}G}$ can be represented as
\begin{equation}
  \overrightarrow{G^{*}G} = \sum_{i\in H^c}s_i'\mathbf{q}_i.
\end{equation}
We show  $s_i' > 0$ for all $i \in H^c$, then we are done. Using Eq. (\ref{eq:qperp}), we have, for each $i \in H^c$, 
\begin{equation}
  \overrightarrow{G^{*}P_i} \cdot \mathbf{q}_i = -a_i^2 < 0.
\end{equation}
This indicates that the vector $\mathbf{q}_i$ is pointing ``outward'' from the simplex $\mathcal{S}(C)$. Since $P_i$ for any $i \in H^c$ and $G$ are on opposite sides of $\mathcal{S}(H)$, it follows that $s_i' > 0$.
\end{proof}
By substituting $\mathbf{n}$ and $\mathbf{q}_i$'s, we have
  \begin{eqnarray}
    \overrightarrow{BG}
  &=& \sum_{i\in H}\left(s_i + S'\frac{A_C^2}{a_i^2}\right)\mathbf{p}_i
  + \sum_{i\in H^c}\left(S'\frac{A_C^2}{a_i^2} - s_i'\right)\mathbf{p}_i\label{eq:BG}
  \end{eqnarray}
  where $S' = \sum_{i\in H^c}s_i'$.
From the coordinates of $G$ (Eq. \ref{eq:G}), we have
\begin{equation}
  \overrightarrow{BG} = \sum_{i=1}^{c}\left(\frac{(1+D_C)A_C^2}{a_i^2} - \frac{b_i}{a_i}\right)\mathbf{p}_i.
\end{equation}
Comparing this equation with Eq. (\ref{eq:BG}) and noting that the vectors $(\mathbf{p}_i)_{i\in C}$ comprise an orthogonal basis set, we have
\begin{eqnarray}
s_i + S'\frac{A_C^2}{a_i^2} &=& \frac{(1+D_C)A_C^2}{a_i^2} - \frac{b_i}{a_i}, ~i\in H,\label{eq:s}\\
S'\frac{A_C^2}{a_i^2} - s_i' &=&\frac{(1+D_C)A_C^2}{a_i^2} - \frac{b_i}{a_i}, ~i\in H^c.\label{eq:sprime}
\end{eqnarray}
By summing Eq. (\ref{eq:sprime}) over $i \in H^c$ and recalling $S' = \sum_{i\in H^c}s_i'$, we have
\begin{equation}
  S' = 1 + D_C - (1 + D_H)\frac{A_H^2}{A_C^2}.
\end{equation}
Substituting this into Eqs. (\ref{eq:s}) and (\ref{eq:sprime}), we have the coefficients $\{s_i\}$ and $\{s_i'\}$:
\begin{eqnarray}
  s_i &=& \frac{(1+D_H)A_H^2}{a_i^2} - \frac{b_i}{a_i}, i \in H;\label{eq:siopt}\\
  s_i' &=& -\frac{(1+D_H)A_H^2}{a_i^2} + \frac{b_i}{a_i}, i \in H^c.
\end{eqnarray}
Thus, $s_i > 0$ and $s_i' \geq 0$ translate into
\begin{eqnarray}
  a_i b_i &< & (1 + D_H)A_H^2, i \in H,\label{eq:abH}\\
  a_i b_i &\geq & (1 + D_H)A_H^2, i \in H^c.\label{eq:abHc}
\end{eqnarray}
Eqs. (\ref{eq:abH}) and (\ref{eq:abHc}) correspond to Eqs. (\ref{eq:ineqdH}) and (\ref{eq:ineqdHc}) in Proposition \ref{prop:H}, respectively.
Note that, given that $H$ is the optimal subset, Eq. (\ref{eq:bgstar}) holds for all the above cases. The coordinates of $G^{*} = (\xi_1^*, \xi_2^*, \cdots, \xi_c^*)$ are given by
\begin{equation}
  \xi_i^* = \left\{
  \begin{array}{cl}
    a_is_i + b_i & (i\in H),\\
    b_i & (i \in H^c).
  \end{array}\right.
\end{equation}
Thus, the optimal membership function is given by
\begin{equation}
  u_i^{*} = \left\{
  \begin{array}{cl}
    s_i & (i \in H),\\
    0 & (i \in H^c).
  \end{array}\right.
\end{equation}
(c.f. Eq. \ref{eq:xi}). Substituting Eq. (\ref{eq:siopt}), we obtain Eq. (\ref{eq:memfunct}) in Proposition \ref{prop:H}.
It is noted from Eq. (\ref{eq:BG}) that
\begin{equation}
  \xi_i^G - b_i = (s_i + S'A_C^2/a_i^2)a_i > 0
\end{equation}
for $i \in H$ because $s_i, S', A_C^2 $ and $a_i$ are  all positive. Thus, we have
\begin{lemma}
  If a vertex $i\in C$ is in the optimal facet $H$, i.e., $i \in H$, then $\xi_i^G > b_i$.
\end{lemma}
Its contrapositive is a useful corollary:
\begin{corollary}\label{cor:H}
  If $\xi_i^G \leq b_i$, then $i \notin H$.
\end{corollary}
Thus, we can eliminate the vertices that do not belong to the optimal subset based on the sign of $\xi_i^G - b_i$ (or $u_i$). 

In the above lemmas, the optimal facet $H$ is assumed to be known. To find the optimal facet, we note the representation of the vector $\overrightarrow{BG}$ in terms of $\{\mathbf{p}_i\}_{i\in H}$ and $\{\mathbf{q}_i\}_{i\in H^c}$ (Eq. \ref{eq:Hrep}) . The following lemma shows that the subset of $C$ admitting such a representation is unique.
\begin{lemma}\label{lemma:facet}
The subset $H \subset C$ with the following properties is unique: 
\begin{equation}
  \overrightarrow{BG} = \sum_{i\in H}s_i \mathbf{p}_i + \sum_{i\in H^c}s_i'\mathbf{q}_i
  \end{equation}
with
\begin{eqnarray}
  s_i & >& 0, i \in H;\\
  \sum_{i\in H}s_i &= & 1;\\
  s'_i & \geq & 0 , i \in H^c
\end{eqnarray}
\end{lemma}
\begin{proof}
Suppose there are two such subsets $H_1$ and $H_2$. Then, we can represent the vector $\overrightarrow{BG}$ in two different ways:
\begin{eqnarray}
  \overrightarrow{BG}
  &=& \sum_{i\in H_1}\left(s_i + S'\frac{A^2}{a_i^2}\right)\mathbf{p}_i
  + \sum_{i\in H_1^c}\left(S'\frac{A^2}{a_i^2} - s_i'\right)\mathbf{p}_i\\
  &=& \sum_{i\in H_2}\left(t_i + T'\frac{A^2}{a_i^2}\right)\mathbf{p}_i
  + \sum_{i\in H_2^c}\left(T'\frac{A^2}{a_i^2} - t_i'\right)\mathbf{p}_i
\end{eqnarray}
where $s_i > 0, s_i' \geq 0, S' = \sum_{i\in H_1^c}s_i'  \geq 0$ and
$t_i > 0, t_i' \geq 0, T' = \sum_{i\in H_2^c}t_i'  \geq 0$. Rearrange the second equality into:
\begin{eqnarray}
  \sum_{i\in H_1\cap H_2}\left(s_i - t_i + [S' - T']\frac{A^2}{a_i^2}\right)\mathbf{p}_i
+ \sum_{i\in H_1 \cap H_2^c}\left(s_i + t_i' + [S' - T']\frac{A^2}{a_i^2}\right)\mathbf{p}_i&&\nonumber\\
+ \sum_{i\in H_1^c \cap H_2}\left(-s_i' - t_i + [S' - T']\frac{A^2}{a_i^2}\right)\mathbf{p}_i
+ \sum_{i\in H_1^c \cap H_2^c}\left(-s_i' + t_i' + [S' - T']\frac{A^2}{a_i^2}\right)\mathbf{p}_i &=& 0\nonumber\\
\end{eqnarray}
Because $(\mathbf{p}_i)$ are an orthogonal basis set, all the coefficients of $\mathbf{p}_i$ must be zero. In particular,
\begin{eqnarray}
  s_i + t_i' + [S' - T']\frac{A^2}{a_i^2} &=& 0, ~~i \in H_1 \cap H_2^c,\\
  -s_i' - t_i + [S' - T']\frac{A^2}{a_i^2} &=& 0, ~~ i \in H_1^c \cap H_2.
\end{eqnarray}
Since $s_i > 0$, we have for $i \in H_1\cap H_2^c$
\begin{equation}
  s_i = -t_i' - [S' - T']A^2/a_i^2 > 0,
\end{equation}
hence
\begin{equation}
  [T' -S']A^2/a_i^2 > t_i'.
\end{equation}
However, we also have $t_i' \geq 0$, which indicates that the left-hand side is positive so that
\begin{equation}
  T' > S'.\label{eq:ts}
\end{equation}
Similarly, for $i \in H_1^c\cap H_2$, we have $t_i > 0$, that is,
\begin{equation}
  t_i = -s_i' + [S' - T']A^2/a_i^2 > 0,
\end{equation}
and hence
\begin{equation}
  [S' - T']A^2/a_i^2 > s_i'.
\end{equation}
But we also have $s_i' \geq 0$, which leads to
\begin{equation}
  S' > T'.\label{eq:st}
\end{equation}
This is a contradiction.
\end{proof}
\begin{corollary}
  The subset $H \subset C$ with the properties in Lemma \ref{lemma:facet} is the optimal subset.
\end{corollary}

\section{Experimental Results}
In this section we provide the results of the proposed HFC method applied to simulated and real-world data, and compare them to conventional methods.

\subsection{Datasets} 
Twenty simulated data generated using the \textit{MixSim} R package \citep{melnykov2012} and five real-world datasets from the UCI repository \citep{dua2019uci}  were used. The dataset specifications detailing the number of clusters $k$, cluster distribution $Pi$, the dimension of data points $p$, the number of data points $N$, and the average overlap of  MixSim datasets are listed in Table \ref{tab:MixSimSpecs} and the UCI datasets Appendicitis, Iris, Seeds, WDBC and Wine  in Table\ref{tab:UciSpecs}.
\begin{table*}[t]
\renewcommand{\arraystretch}{1.2}
\centering
\caption{MixSim generated dataset specifications. \label{tab:MixSimSpecs}}
{\begin{tabular}{l l l l l l}
\hline\noalign{\smallskip}
Dataset	&k$^a$	&Pi$^b$	&p$^c$	&N$^d$	&overlap$^e$\\		
\noalign{\smallskip}
\hline
Sim2	&2	&0.5, 0.5	        &2	&200	&0.001\\
Sim3	&2	&0.5, 0.5	        &2	&200	&0.01\\
Sim6	&2	&0.5, 0.5        	&2	&200	&0.05\\
\hline
Sim9	&2	&0.9, 0.1       	&2	&200	&0.0005\\
Sim14	&2	&0.9, 0.1       	&2	&30	    &0.001\\
Sim8	&2	&0.9, 0.1       	&2	&200	&0.01\\
Sim7	&2	&0.9, 0.1       	&2	&200	&0.05\\
\hline
Sim10	&2	&0.8, 0.2       	&2	&200	&0.0001\\
Sim17	&2	&0.8, 0.2	        &2	&50	    &0.005\\
Sim1	&2	&0.2, 0.8	        &2	&200	&0.001\\
Sim4	&2	&0.2, 0.8        	&2	&200	&0.01\\
Sim5	&2	&0.2, 0.8	        &2	&200	&0.05\\
\hline
Sim13	&2	&0.7, 0.3       	&2	&30	    &0.001\\
Sim18	&2	&0.7, 0.3	        &2	&50 	&0.005\\
Sim11	&2	&0.7, 0.3       	&2	&200	&0.01\\
Sim12	&2	&0.7, 0.3       	&2	&200	&0.05\\
\hline
Sim19	&3	&0.8, 0.1, 0.1	    &2	&50	    &0.005\\
\hline
Sim15	&3	&0.7, 0.2, 0.1    	&2	&50 	&0\\
Sim20	&3	&0.7, 0.2, 0.1 	    &2	&50 	&0.005\\
\hline
Sim16	&4	&0.6, 0.2, 0.1, 0.1	&2	&50 	&0\\
\hline 
\end{tabular}}\\
$^a$Number of clusters.
$^b$Cluster distribution.
$^c$Data dimension.
$^d$Number of data points.
$^e$Average overlap (BarOmega).
\end{table*}

\begin{table*}[t]
\renewcommand{\arraystretch}{1.2}
\centering
\caption{UCI dataset specifications. \label{tab:UciSpecs}}
{\begin{tabular}{l l l l l}
\hline\noalign{\smallskip}
Dataset	        &k$^a$	&Pi$^b$	                    &p$^c$	&N$^d$\\		
\noalign{\smallskip}
\hline
Appendicitis	&2	&0.198, 0.802	        &9	&106\\
Iris	        &3	&0.333, 0.333, 0.333	&4	&150\\
Seeds	        &3	&0.333, 0.333, 0.333	&7	&210\\
WDBC	        &2	&0.373, 0.627	        &32	&569\\
Wine	        &3	&0.331, 0.399, 0.270	&13	&178\\
\hline 
\end{tabular}}\\
$^a$Number of clusters.
$^b$Cluster distribution.
$^c$Data dimension.
$^d$Number of data points.
\end{table*}

\subsection{Evaluation measures}
Four evaluation measures were recorded, external measure Corrected Rand Index or Adjusted Rand Index (ARI) \citep{gordon1999classification} and internal measures entropy of the distribution of cluster memberships (Entropy) \citep{meilua2007comparing}, separation index (SIndex) and Dunn2 index \citep{halkidi01} based on the \textit{fpc} R package \citep{hennig2018fpc}. 

\textit{Corrected Rand Index} assessed the similarity between the clustering result and the ground truth of the same set of objects, adjusted for chance. Values of near 1 indicate high similarity. \textit{Entropy} distribution of cluster memberships is calculated based on the negative of the sum of product of probability and log probability of the data points belonging to a cluster, across all clusters. A near zero value indicate little uncertainty in the structure of the cluster. The \textit{separation index} is based on the distances of every point to the closest point not in the same cluster. Larger values indicate distinctiveness of cluster. The results of the separation index are found in the supplementary material. \textit{Dunn2} calculates the ratio of minimum average dissimilarity between two clusters and the maximum average within cluster dissimilarity, where a larger value indicates more distinct clusters. 

We also applied the \textit{plotcluster} function from \textit{fpc} R package to plot the data points and the cluster they belong to (indicated by the numbered labels while the colors ground truth).

\subsection{Clustering algorithms}
We compared the performance of our proposed hybrid fuzzy-crisp HFC algorithm with two conventional clustering algorithms, K-means (KM) and Fuzzy c-means (FCM) to investigate in the behaviors of the three algorithms in handling datasets with imbalanced clusters.

\subsection{Experiment settings}
The experiments were repeated 30 times with each run having a maximum number of iterations set at 50 and the algorithm stops when the performance by a threshold of less than 0.0001. $\alpha$ is initialized as $1/c$ and updated at each iteration but $\beta$ is kept fixed at $1/c$ throughout all iterations.

 The purpose of this study is not to investigate in cluster initialization, but the behavior of the clustering algorithms. To initialize the cluster centers and avoid bad start, two data points from the same clusters are chosen for each cluster and averaged. The same cluster centers are used to initialize the three algorithms HFC, KM and FCM. 

For the synthetic datasets generated using MixSim, a different dataset is generated for each of the 30 runs, each with the same settings as detailed in Table \ref{tab:MixSimSpecs}. The reason we did this is to see how well the algorithms perform on different datasets with same settings since on the same dataset, the performance usually do not vary much across different runs. This was also observed on UCI datasets by their low standard deviation in Table \ref{tab:UCIResults}. The average ARI, Entropy and Dunn2 results and their standard deviation are presented. For full results with Sindex measure and their best result, please refer to the supplementary materials.

%% \section{Empirical Results}
%% Tables \ref{tab:SimDiffResults0.50.5}-\ref{tab:SimDiffResults0.60.20.10.1} show the performances of the algorithms FC, KM and FCM on MixSim datasets groupby \textit{vector of mixing proportion} (cluster distribution) \citep{melnykov2012} and in ascending order of Mix Sim parameter BarOmega (average overlap) values indicated in brackets. This is done to study whether the effect of overlapping data points on the clustering performance. Table \ref{tab:UCIResults} shows the performances of the three algorithms on the five UCI datasets. The mean, standard deviation and best results are presented, with the best Entropy values having minimum values while maximum for ARI, Sindex and Dunn2. Figure \ref{fig:0.50.5} - ???? shows the cluster plot of clustering solutions from a selected run to provide visual examination of the solutions found by the algorithms.

\subsection{MixSim datasets}
MixSim is an R package \citep{melnykov2012} that generates random points from Gaussian mixtures. We generated various datasets with two to four clusters with different cluster sizes (Table \ref{tab:MixSimSpecs}).

\begin{figure}[t]
\centering
  % \widefigure
% \hspace{-8pt}
\includegraphics[width=1.0\textwidth]{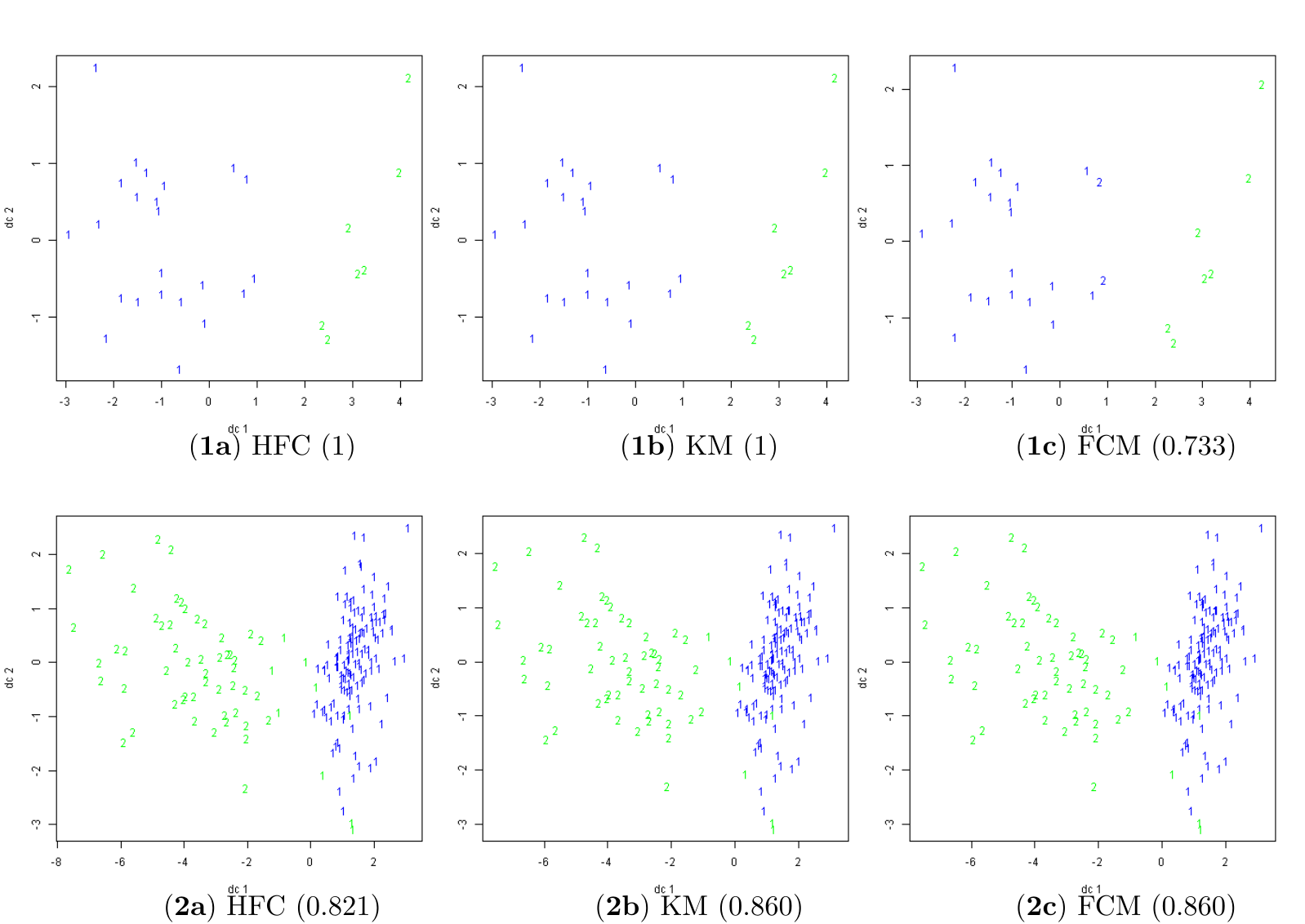}
\caption{Cluster plots  of one run on 1) Sim2 (overlap of 0.001) 2) Sim3 (overlap of 0.01) and 3) Sim6 (overlap of 0.05) with cluster distribution (0.5, 0.5). The ARI values are reported in parentheses. See also Table \ref{tab:SimDiffResults0.50.5}. \label{fig:0.50.5}}
\end{figure}

\paragraph{Balanced datasets}
Table \ref{tab:SimDiffResults0.50.5} shows the performance of HFC, KM and FCM on balanced MixSim datasets with cluster distribution of (0.5, 0.5). HFC outperforms the other algorithms in terms of ARI measure on the Sim2 data which has the least overlap while KM performs best on Sim3 and Sim6.  HFC is observed to produce clusters with the best Entropy in all three datasets.
%Discussion
We observed that for balanced dataset where there is little or no overlap, HFC can be as competitive as KM and FCM, if not outperforming them as shown in Figure \ref{fig:0.50.5} (1a) where more data points at the edge of the green cluster are correctly assigned by HFC than KM (in 1b) and FCM (in 1c). Where there is more overlap in Figure \ref{fig:0.50.5} (3a), HFC is unable to distinguish clusters for data points sitting at the boundaries where the overlap is occurring.

\begin{figure}[t]
\centering
  % \widefigure
% \hspace{-8pt}
\includegraphics[width=1.0\textwidth]{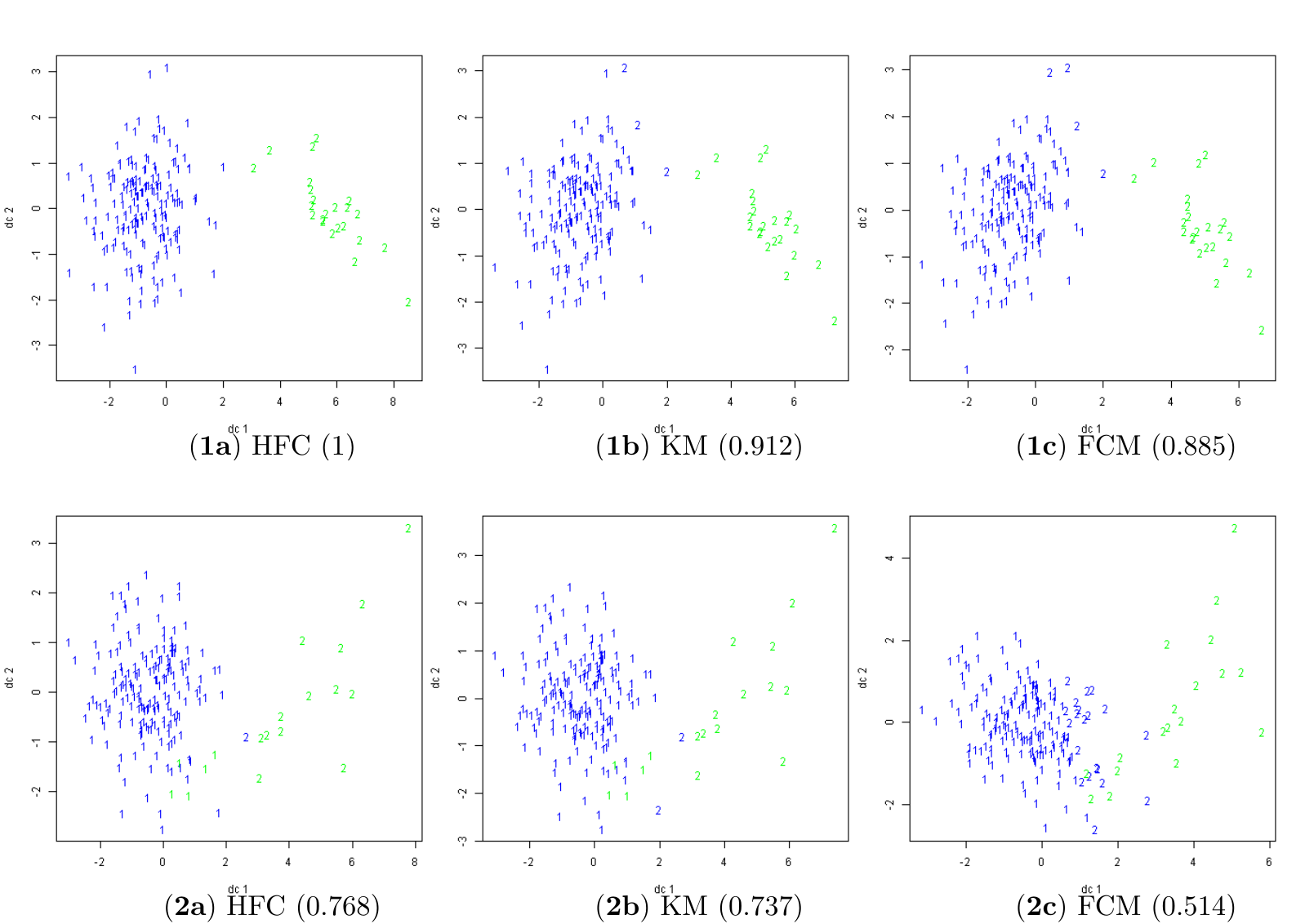}
\caption{\label{fig:0.90.1}Cluster plots  of one run on 1) Sim9 (overlap of 0.0005) and 2) Sim7 (overlap of 0.05) with cluster distribution (0.9, 0.1). The ARI values are reported in parentheses. See also Table \ref{tab:SimDiffResults0.90.1}.}
\end{figure}

\begin{figure}[t]
\centering
  % \widefigure
% \hspace{-8pt}
\includegraphics[width=1.0\textwidth]{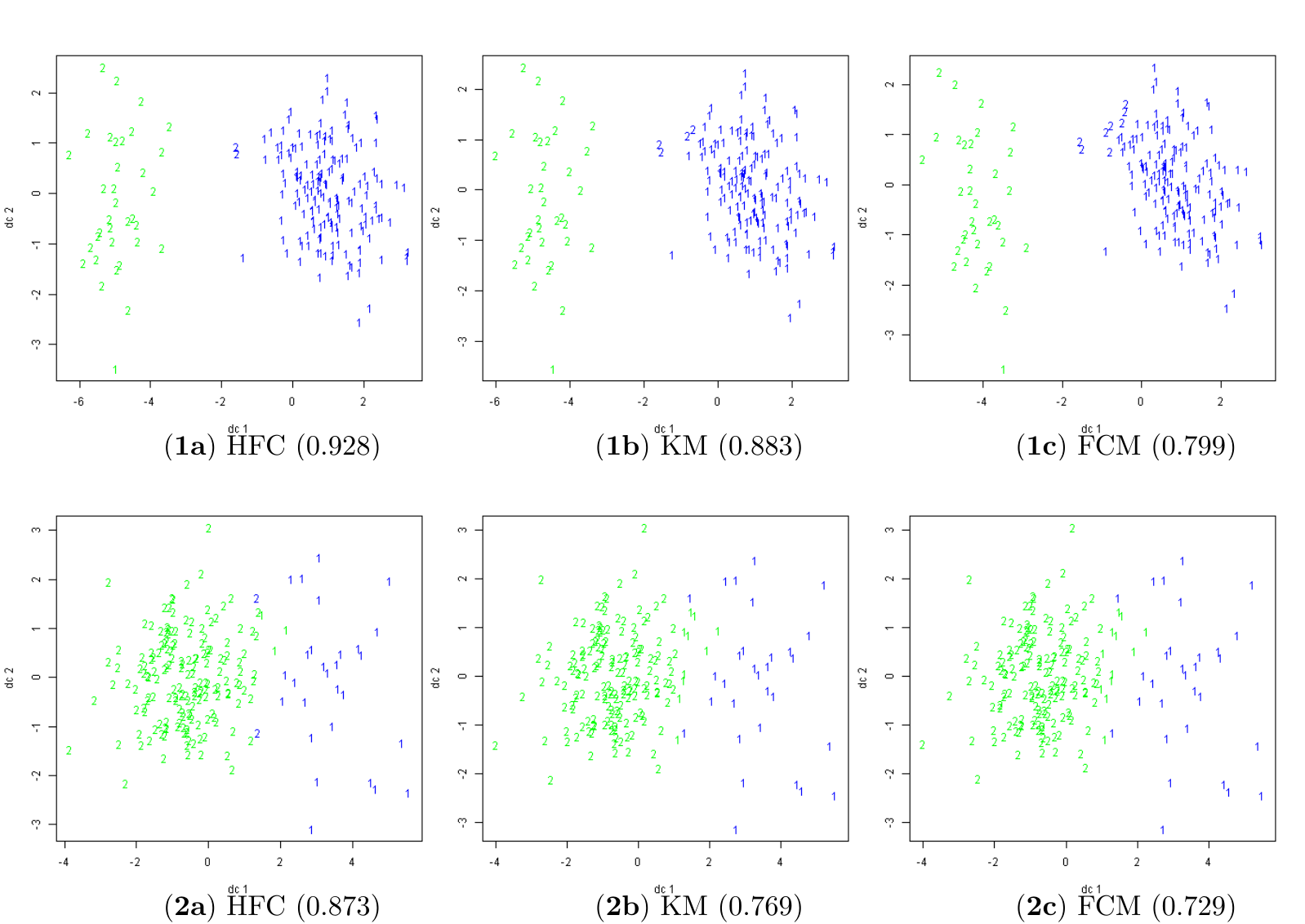}
\vspace{-12pt}
\caption{\label{fig:0.80.2}Cluster plots  of one run on 1) Sim10 (overlap of 0.0001) and 2) Sim5 (overlap of 0.05) with cluster distribution (0.8, 0.2). The ARI values are reported in parentheses. See also Table \ref{tab:SimDiffResults0.80.2}.}
\end{figure}

%Table 4
\paragraph{Imbalanced datasets}
Table \ref{tab:SimDiffResults0.90.1} shows the performance of HFC, KM and FCM on imbalanced MixSim datasets with cluster distribution of (0.9, 0.1). HFC performs the best out of all algorithms in terms of ARI, Entropy and Sindex measures on the Sim9, Sim14, Sim8 and Sim7 data which have increasing amount of overlap from 0.0005 for Sim9 to 0.05 for Sim7. As higher degree of overlap is introduced, the performance decrease for all algorithms, with HFC performing the best. This demonstrates that HFC outperforms K and FCM when the dataset is highly imbalanced. Figure \ref{fig:0.90.1} shows the clustering results on datasets of the two degrees of overlapping, Sim9 with very slight overlap of 0.0005 and Sim7 with slight overlap of 0.05. In both datasets, we observed visually that HFC were able to cluster slightly better than KM while FCM performed much worse than the other two algorithms. Similar trends were found using simulated datasets with cluster distribution of (0.8, 0.2) in Table \ref{tab:SimDiffResults0.80.2} and Figure \ref{fig:0.80.2}.

\begin{figure}[t]
\centering
  % \widefigure
% \hspace{-8pt}
\includegraphics[width=1.0\textwidth]{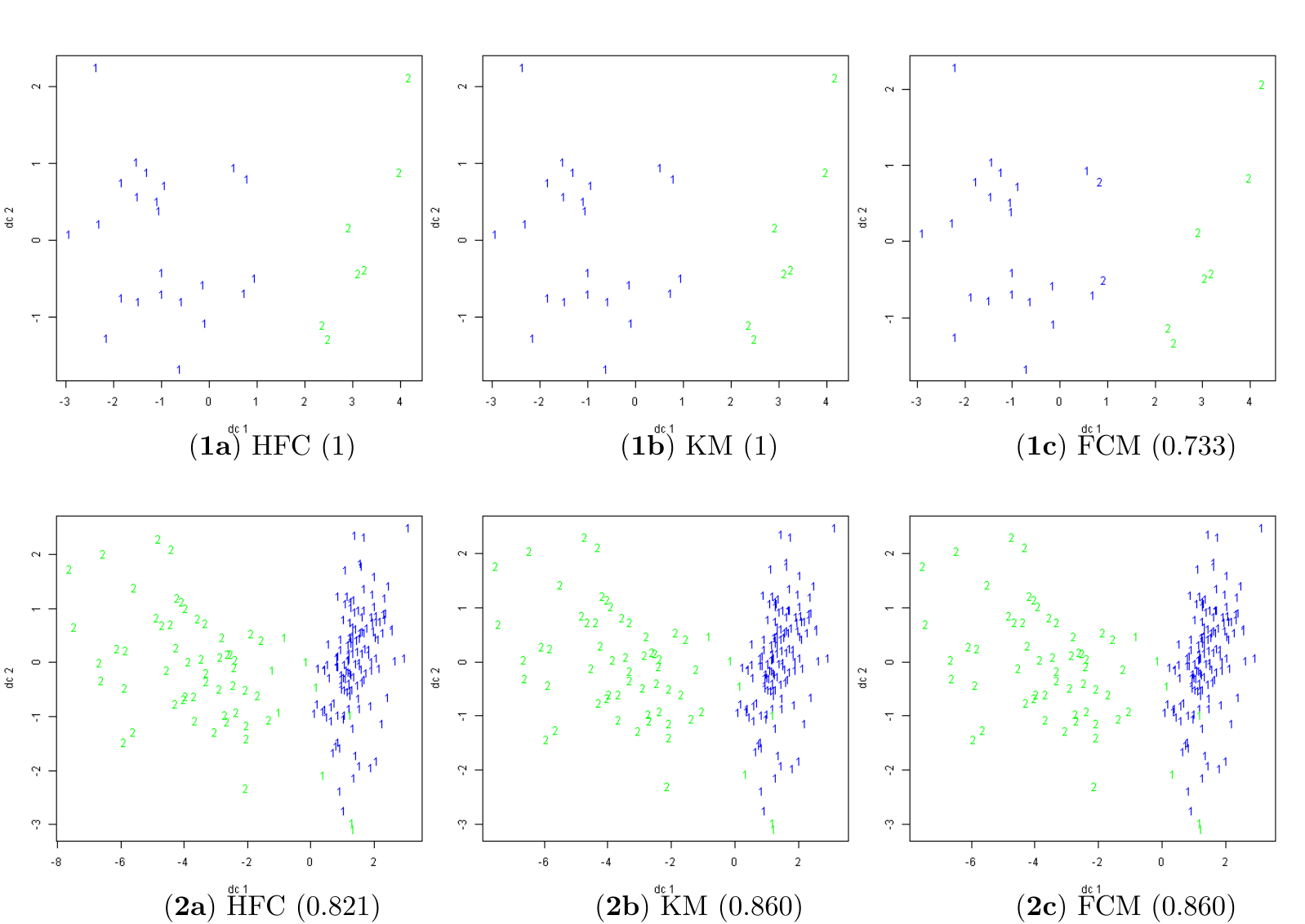}
\vspace{-12pt}
\caption{\label{fig:0.70.3}Cluster plots  of one run on 1) Sim13 (overlap of 0.001) and 2) Sim12 (overlap of 0.05) with cluster distribution (0.7, 0.3). The ARI values are reported in parentheses. See also Table \ref{tab:SimDiffResults0.70.3}. }
\end{figure}

%Table 6, Fig 4
In Table \ref{tab:SimDiffResults0.70.3} and Figure \ref{fig:0.70.3}, we observed that HFC can still outperform KM and FCM in datasets with cluster distribution (0.7, 0.3). However, on Sim12 data with slightly more overlap, KM and FCM outperforms HFC. This demonstrate that HFC can handle imbalanced dataset with very slight overlapping ($< 0.01$).

\begin{figure}[t]
\centering
  % \widefigure
% \hspace{-8pt}
\includegraphics[width=1.0\textwidth]{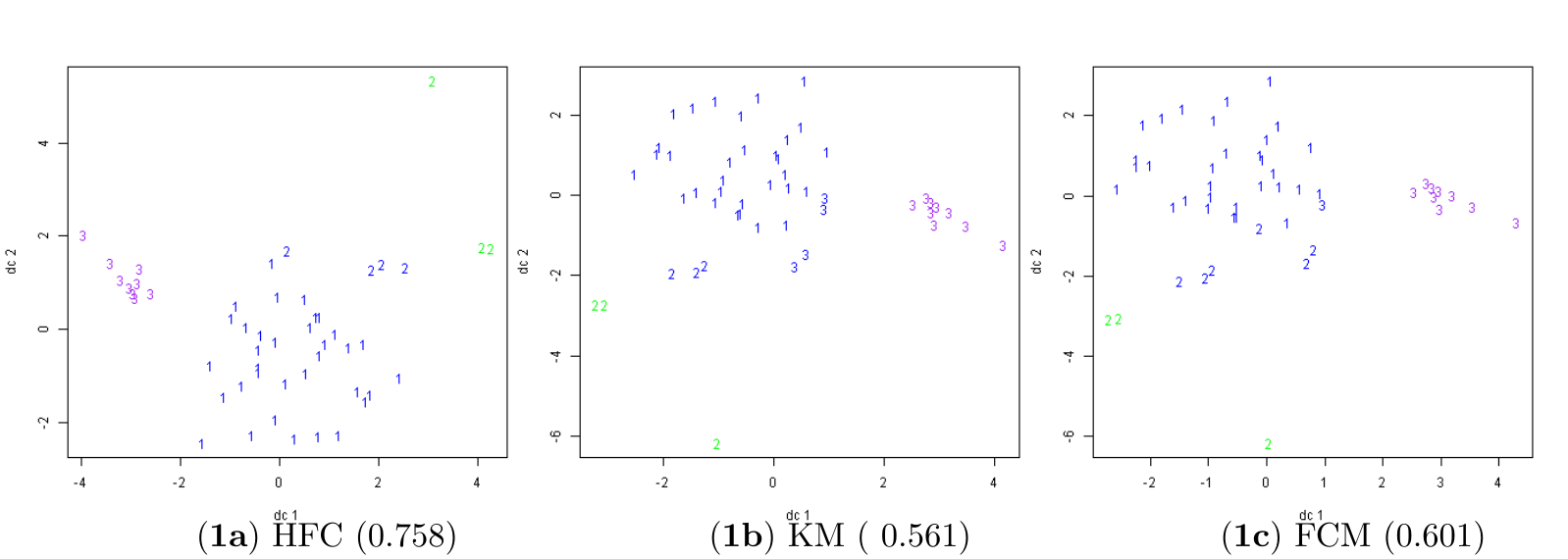}
\caption{\label{fig:0.80.10.1}Cluster plots  of one run on 1) Sim19 (overlap of 0.005) with cluster distribution (0.8, 0.1, 0.1). The ARI values are reported in parentheses. See also Table \ref{tab:SimDiffResults0.80.10.1}.}
\end{figure}

%Table 7, Fig 5
In Table \ref{tab:SimDiffResults0.80.10.1} and Figure \ref{fig:0.80.10.1}, we observed the results on Sim19 with cluster distribution of (0.8, 0.1, 0.1) and slight overlap where HFC outperformed KM and FCM on average across 30 such datasets on all measures, achieving the maximum average ARI, Sindex and Dunn2 measures and minimum average Entropy.  As there is great disparity of cluster ratios of 0.8 and 0.1, we observed in \ref{fig:0.80.10.1} that KM and FCM wrongly assign labels 2 and 3 to the blue cluster while HFC wrongly labels a few data points of the blue cluster to 2.  

\begin{figure}[t]
\centering
  % \widefigure
% \hspace{-8pt}
\includegraphics[width=1.0\textwidth]{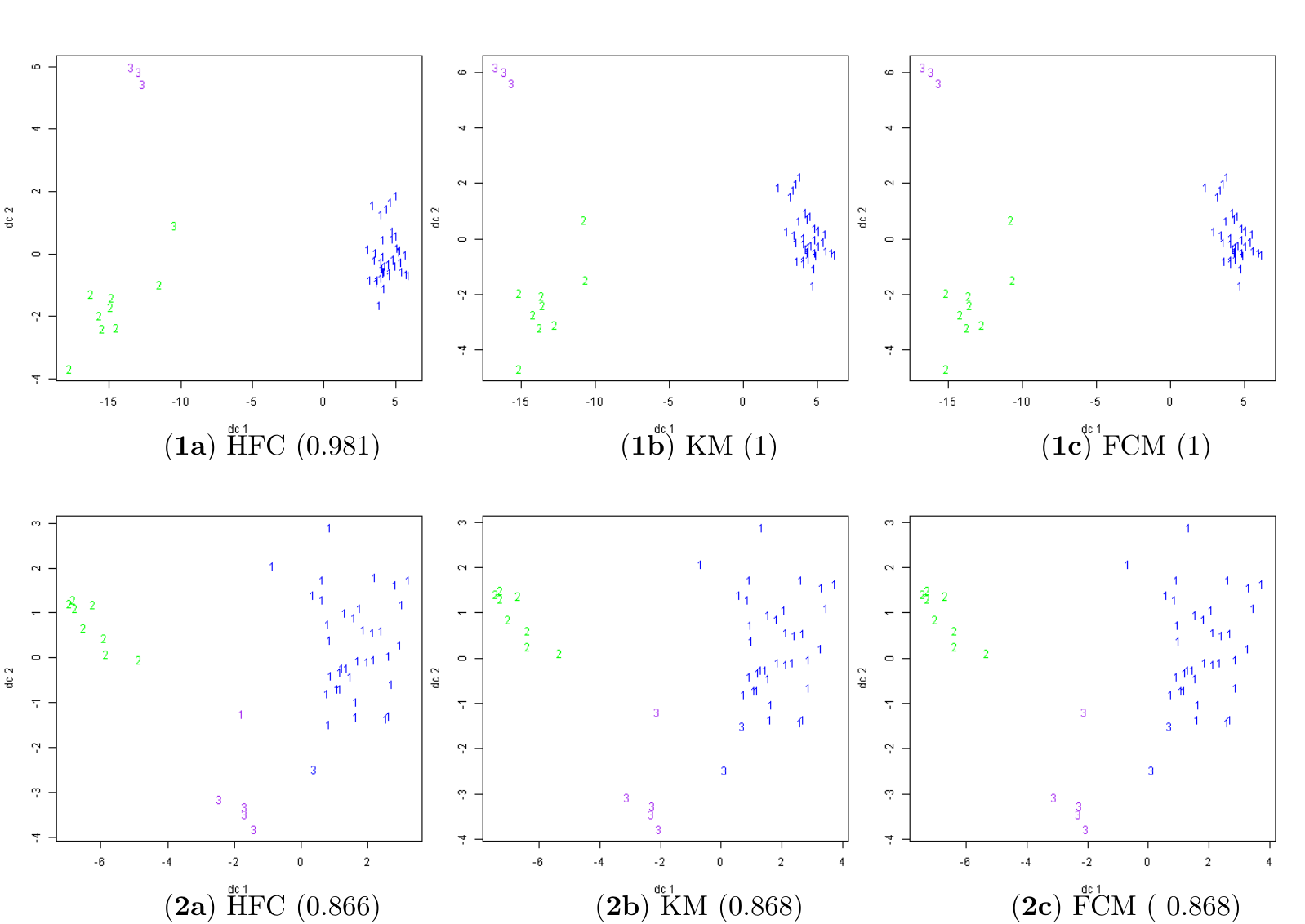}
\caption{\label{fig:0.70.20.1}Cluster plots  of one run on 1) Sim15 (overlap of 0) and 2) Sim20 (overlap of 0.005) with cluster distribution (0.7, 0.2, 0.1). The ARI values are reported in parentheses. See also Table \ref{tab:SimDiffResults0.70.20.1}.}
\end{figure}

%Table 8, Fig 6
Table \ref{tab:SimDiffResults0.70.20.1} and Figure  \ref{fig:0.70.20.1} show results on imbalanced datasets with 3 clusters and cluster distribution of (0.7, 0.2, 0.1). KM and FCM outperformed HFC in Sim15 which has no overlap while KM outperformed HFC in Sim20 which has 0.005 overlap. This may indicate that as disparity of cluster distribution reduces, KM and FCM will outperform HFC. As the number of cluster increases, there may be a tendency for cluster distribution differences to be smaller such that KM and FCM will outperform HFC, as shown in Table \ref{tab:SimDiffResults0.60.20.10.1} and Figure  \ref{fig:0.60.20.10.1}. %Table 9, Fig 7

\begin{landscape} 
\begin{table*}[t]
\renewcommand{\arraystretch}{1.1}
\centering
\caption{Evaluation on unique MixSim datasets with Pi = (0.5, 0.5) (the average overlap value indicated in parentheses). See also Figure \ref{fig:0.50.5}. \label{tab:SimDiffResults0.50.5}}%\setlength{\tabcolsep}{2pt} 
{\begin{tabular}{l c c c c c c c c c c c}
\setlength{\tabcolsep}{3pt} 
\\ \hline\noalign{\smallskip}
	&	& \multicolumn{3}{c}{ARI}			& \multicolumn{3}{c}{Entropy}			&  \multicolumn{3}{c}{Dunn2}\\		
Dataset	&	&HFC	&KM	&FCM	&HFC	&KM	&FCM	&HFC	&KM	&FCM\\
\noalign{\smallskip}
\hline
Sim2	&Mean	&\textbf{0.831}	&0.772	&0.73	
                &\textbf{0.538}	&0.561	&0.573	%entropy
                &\textbf{2.14}	&2.099	&2.079\\
	
(0.001)	&StDev	&0.264	&0.285	&0.296	
	            &0.068	&0.076	&0.075	
	            &0.421	&0.428	&0.443\\
\\[0.5pt]

Sim3	&Mean	&0.798	&\textbf{0.804}	&0.724	
                &\textbf{0.516}	&0.527	&0.554	%entropy
                &\textbf{2.104}	&2.077	&1.983\\

(0.01)    &StDev	&0.257	&0.244	&0.28	
                    &0.087	&0.076	&0.084	
                    &0.549	&0.548	&0.56\\
\\[0.5pt]

Sim6	&Mean	&0.673	&\textbf{0.684}	&0.682	
                &\textbf{0.675}	&0.678	&0.681	%entropy
                &1.747	&\textbf{1.756}	&1.749\\
	
(0.05)	&StDev	&0.18	&0.173	&0.149	
	            &0.023	&0.017	&0.014	
	            &0.276	&0.268	&0.27\\
\hline 
\end{tabular}}
\end{table*}
\end{landscape}

\begin{landscape} 
\begin{table*}[t]
\renewcommand{\arraystretch}{1.2}
\centering
\caption{\label{tab:SimDiffResults0.90.1}Evaluation on unique MixSim datasets with Pi = (0.9, 0.1) (the average overlap value indicated in parentheses). See also Figure \ref{fig:0.90.1}}%\setlength{\tabcolsep}{2pt} 
{\begin{tabular}{l c c c c c c c c c c c}
\renewcommand{\arraystretch}{1.2}
\setlength{\tabcolsep}{3pt} 
\\ \hline\noalign{\smallskip}
	&	& \multicolumn{3}{c}{ARI}			& \multicolumn{3}{c}{Entropy}			& \multicolumn{3}{c}{Dunn2}\\		
Dataset	&	&HFC	&KM	&FCM	&HFC	&KM	&FCM	&HFC	&KM	&FCM	\\
\noalign{\smallskip}
\hline
Sim9	    &Mean	&\textbf{0.963}	&0.959	&0.849	
            &\textbf{0.342}	&0.347	&0.391	%entropy
                    &\textbf{2.764}	&2.721	&2.632\\
	
(0.0005)	&StDev	&0.131	&0.132	&0.318	
	        &0.063	&0.061	&0.118	
	        &0.817	&0.847	&0.894\\
\\[0.5pt]
	
Sim14	&Mean	&\textbf{0.851}	&0.831	&0.681	
                &\textbf{0.433}	&0.438	&0.502	%entropy
                &2.302	&\textbf{2.306}	&2.106\\
	
(0.001)	&StDev	&0.319	&0.323	&0.42	
	                &0.122	&0.126	&0.137	
	                &0.745	&0.742	&0.798\\
\\[0.5pt]

Sim8	&Mean	&\textbf{0.797}	&0.715	&0.57	
               &\textbf{0.376}	&0.419	&0.48	%entropy
                &\textbf{2.358}	&2.215	&2.033\\
	
(0.01)	&StDev	&0.329	&0.371	&0.435	
	            &0.139	&0.156	&0.172	
	            &0.568	&0.611	&0.593\\
\\[0.5pt]
	
Sim7	&Mean	&\textbf{0.465}	&0.403	&0.3	
                &\textbf{0.509}	&0.541	&0.594	%entropy
                &\textbf{1.828}	&1.765	&1.656\\
	
(0.05)	&StDev	&0.391	&0.394	&0.362	
	            &0.173	&0.163	&0.14	
	            &0.384	&0.304	&0.278\\
\hline 
\end{tabular}}
\end{table*}
\end{landscape}

\begin{landscape} 
\begin{table*}[t]
\renewcommand{\arraystretch}{1.2}
\centering
\caption{\label{tab:SimDiffResults0.80.2}Evaluation on unique MixSim datasets with Pi = (0.8, 0.2) (the average overlap value indicated in parentheses). See also Figure \ref{fig:0.80.2}.}
{\begin{tabular}{l c c c c c c c c c c c}
\setlength{\tabcolsep}{3pt} 
\\ \hline\noalign{\smallskip}
	&	& \multicolumn{3}{c}{ARI}			& \multicolumn{3}{c}{Entropy}			&  \multicolumn{3}{c}{Dunn2}\\		
Dataset	&	&HFC	&KM	&FCM	&HFC	&KM	&FCM	&HFC	&KM	&FCM	\\
\noalign{\smallskip}
\hline
Sim10	&Mean	&\textbf{0.967}	&0.952	&0.943	
                &\textbf{0.504}	&0.51	&0.512	%entropy
                &\textbf{2.581}	&2.546	&2.542\\
	
(0.0001)	&StDev	&0.091	&0.106	&0.105	
	                &0.048	&0.048	&0.049	
	                &0.642	&0.652	&0.656\\
\\[0.5pt]
	
Sim17	&Mean	&\textbf{0.94}	&0.903	&0.869	
                &\textbf{0.506}	&0.516	&0.53	%entropy
                &\textbf{2.206}	&2.204	&2.149\\
	
(0.005)	&StDev	&0.105	&0.135	&0.179	
	            &0.063	&0.07	&0.074	
	            &0.495	&0.489	&0.49\\
\\[0.5pt]
	
Sim1	&Mean	&\textbf{0.87}	&0.816	&0.797	
                &\textbf{0.517}	&0.536	&0.547	%entropy
                &\textbf{2.236}	&2.189	&2.157\\
                
(0.001)	    &StDev	&0.17	&0.227	&0.21	
	                &0.053	&0.064	&0.063	
	                &0.591	&0.605	&0.62\\
\\[0.5pt]
	
Sim4	&Mean	&\textbf{0.871}	&0.841	&0.808	
                &\textbf{0.512}	&0.523	&0.545	%entropy
                &\textbf{2.131}	&2.092	&2.005\\

(0.01)	    &StDev	&0.188	&0.198	&0.231	
	                &0.064	&0.068	&0.062	
	                &0.512	&0.497	&0.526\\
\\[0.5pt]
	
Sim5	&Mean	&\textbf{0.645}	&0.616	&0.611	
                &\textbf{0.521}	&0.562	&0.582	%entropy
                &\textbf{1.837}	&1.727	&1.672\\
	
(0.05)	&StDev	&0.285	&0.299	&0.301	
	            &0.095	&0.086	&0.078	
	            &0.291	&0.314	&0.3\\
\hline 
\end{tabular}}
\end{table*}
\end{landscape}

\begin{landscape} 
\begin{table*}[t]
\renewcommand{\arraystretch}{1.2}
\centering
\caption{\label{tab:SimDiffResults0.70.3}Evaluation on unique MixSim datasets with Pi = (0.7, 0.3) (the average overlap value indicated in parentheses). See also Figure \ref{fig:0.70.3}.}
{\begin{tabular}{l c c c c c c c c c c c}
\setlength{\tabcolsep}{3pt} 
\\ \hline\noalign{\smallskip}
	&	& \multicolumn{3}{c}{ARI}			& \multicolumn{3}{c}{Entropy}			&  \multicolumn{3}{c}{Dunn2}\\		
Dataset	&	&HFC	&KM	&FCM	&HFC	&KM	&FCM	&HFC	&KM	&FCM	\\
\noalign{\smallskip}
\hline
Sim13	&Mean	&\textbf{0.91}	&0.89	&0.886	
                &\textbf{0.602}	&0.614	&0.613	%entropy
                &2.284	&2.289	&\textbf{2.291}\\
	
(0.001)	&StDev	&0.194	&0.231	&0.214	
	            &0.088	&0.081	&0.083	
	            &0.661	&0.662	&0.656\\
\\[0.5pt]
	
Sim18	&Mean	&\textbf{0.862}	&0.859	&0.817	
                &\textbf{0.593}	&0.605	&0.607 %entropy	
                &2.084	&2.033	&2.055\\
	
(0.005)	&StDev	&0.154	&0.164	&0.208	
	            &0.059	&0.053	&0.054	
	            &0.557	&0.546	&0.552\\
\\[0.5pt]	
	
Sim11	&Mean	&\textbf{0.803}	&0.785	&0.768	
                &\textbf{0.615}	&0.619	&0.633	%entropy
                &1.927	&\textbf{1.957}	&1.917\\
	
(0.01)	&StDev	&0.237	&0.24	&0.223	
	            &0.047	&0.048	&0.045	
	            &0.409	&0.38	&0.41\\
\\[0.5pt]

Sim12	&Mean	&0.631	&\textbf{0.673}	&0.658	
                &\textbf{0.598}	&0.61	&0.624	%entropy
                &\textbf{1.934}	&1.913	&1.877\\
	
(0.05)	&StDev	&0.264	&0.235	&0.231	
	            &0.078	&0.073	&0.057	
	            &0.339	&0.33	&0.347\\
\hline 
\end{tabular}}
\end{table*}
\end{landscape}

\begin{landscape} 
\begin{table*}[t]
\renewcommand{\arraystretch}{1.2}
\centering
\caption{\label{tab:SimDiffResults0.80.10.1}Evaluation on unique MixSim datasets with Pi = (0.8, 0.1, 0.1) (the average overlap value indicated in parentheses). See also Figure \ref{fig:0.80.10.1}.}
\vspace{-10pt}
{\begin{tabular}{l c c c c c c c c c c c}
\setlength{\tabcolsep}{3pt} 
\\ \hline\noalign{\smallskip}
	&	& \multicolumn{3}{c}{ARI}			& \multicolumn{3}{c}{Entropy}			&  \multicolumn{3}{c}{Dunn2}\\		
Dataset	&	&HFC	&KM	&FCM	&HFC	&KM	&FCM	&HFC	&KM	&FCM	\\
\noalign{\smallskip}
\hline
Sim19	&Mean	&\textbf{0.879}	&0.872	&0.817	
                &\textbf{0.7}	&0.708	&0.741	%entropy
                &\textbf{2.348}	&2.342	&2.246\\
                
(0.005)	&StDev	&0.21	&0.215	&0.273	
	            &0.141	&0.142	&0.162	
	            &0.694	&0.679	&0.734\\
\hline 
\end{tabular}}
\end{table*}
 
\begin{table*}[t]
\renewcommand{\arraystretch}{1.2}
\centering
\caption{\label{tab:SimDiffResults0.70.20.1}Evaluation on unique MixSim datasets with Pi = (0.7, 0.2, 0.1) (the average overlap value indicated in parentheses). See also Figure \ref{fig:0.70.20.1}.}
\vspace{-10pt}
{\begin{tabular}{l c c c c c c c c c c c}
\setlength{\tabcolsep}{3pt} 
\\ \hline\noalign{\smallskip}
	&	& \multicolumn{3}{c}{ARI}			& \multicolumn{3}{c}{Entropy}			&  \multicolumn{3}{c}{Dunn2}\\		
Dataset	&	&HFC	&KM	&FCM	&HFC	&KM	&FCM	&HFC	&KM	&FCM	\\
\noalign{\smallskip}
\hline
Sim15	&Mean	&0.929	&0.932	&\textbf{0.933}	
                &\textbf{0.849}	&0.854	&0.855	%entropy
                &\textbf{4.412}	&4.409	&4.41\\
	
(0)	&StDev	&0.211	&0.202	&0.199	
	        &0.09	&0.1	&0.101	
	        &1.781	&1.785	&1.784\\
\\[0.5pt]
Sim20	&Mean	&0.824	&\textbf{0.839}	&0.788	
                &\textbf{0.846}	&0.869	&0.882	%entropy
                &\textbf{1.997}	&1.972	&1.958\\
	
(0.005)	&StDev	&0.218	&0.194	&0.242	
	            &0.12	&0.111	&0.117	
	            &0.681	&0.675	&0.661\\
\hline 
\end{tabular}}
\end{table*}
\end{landscape}

\begin{landscape} 
\begin{table*}[t]
\renewcommand{\arraystretch}{1.2}
\centering
\caption{\label{tab:SimDiffResults0.60.20.10.1}Evaluation on unique MixSim datasets with Pi = (0.6, 0.2, 0.1, 0.1) (the average overlap value indicated in parentheses). See also Figure \ref{fig:0.60.20.10.1}.}
\vspace{-10pt}
{\begin{tabular}{l c c c c c c c c c c c}
\setlength{\tabcolsep}{3pt} 
\\ \hline\noalign{\smallskip}
	&	& \multicolumn{3}{c}{ARI}			& \multicolumn{3}{c}{Entropy}			&  \multicolumn{3}{c}{Dunn2}\\		
Dataset	&	&HFC	&KM	&FCM	&HFC	&KM	&FCM	&HFC	&KM	&FCM	\\
\noalign{\smallskip}
\hline
Sim16	&Mean	&0.996	&1	    &1	
                &1.077	&1.076	&1.076	%entropy
                &3.956	&4.054	&4.054\\
	
(0)	&StDev	&0.018	&0	    &0	
	        &0.104	&0.104	&0.104	
	        &1.71	&1.606	&1.606\\
\hline 
\end{tabular}}
\end{table*}

\begin{figure}[t]
\centering
  % \widefigure
% \hspace{-8pt}
\includegraphics[width=1.1\textwidth]{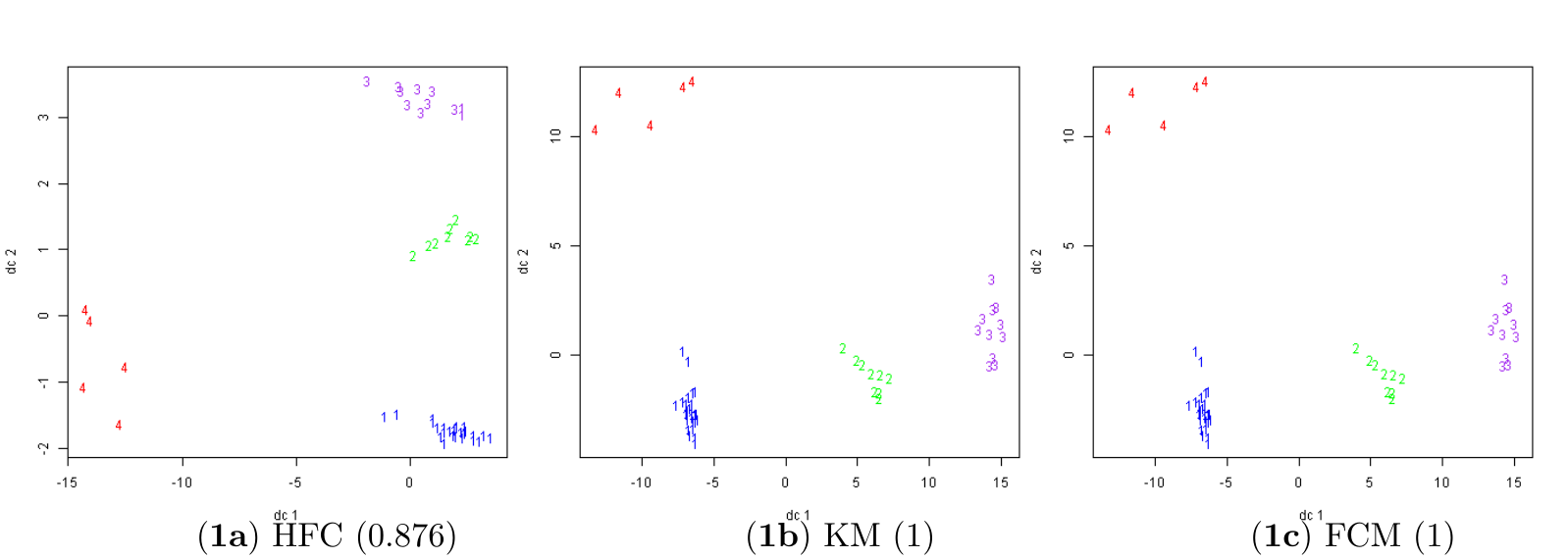}
\vspace{-12pt}
\caption{\label{fig:0.60.20.10.1}Cluster plots  of one run on 1) Sim16 (overlap of 0) with cluster distribution (0.6, 0.2, 0.1, 0.1). The ARI values are reported in parentheses. See also Table \ref{tab:SimDiffResults0.60.20.10.1}.}
\end{figure} 

\end{landscape}

\subsection{Real-world data}
In this section, we present and analyze the results from testing selected real-world datasets. In Table \ref{tab:UCIResults}, HFC outperformed KM and FCM on the Appendicitis, WDBC and Seeds datasets in terms of ARI measures. 

\begin{figure}[t]
\centering
  % \widefigure
% \hspace{-8pt}
\includegraphics[width=1.0\textwidth]{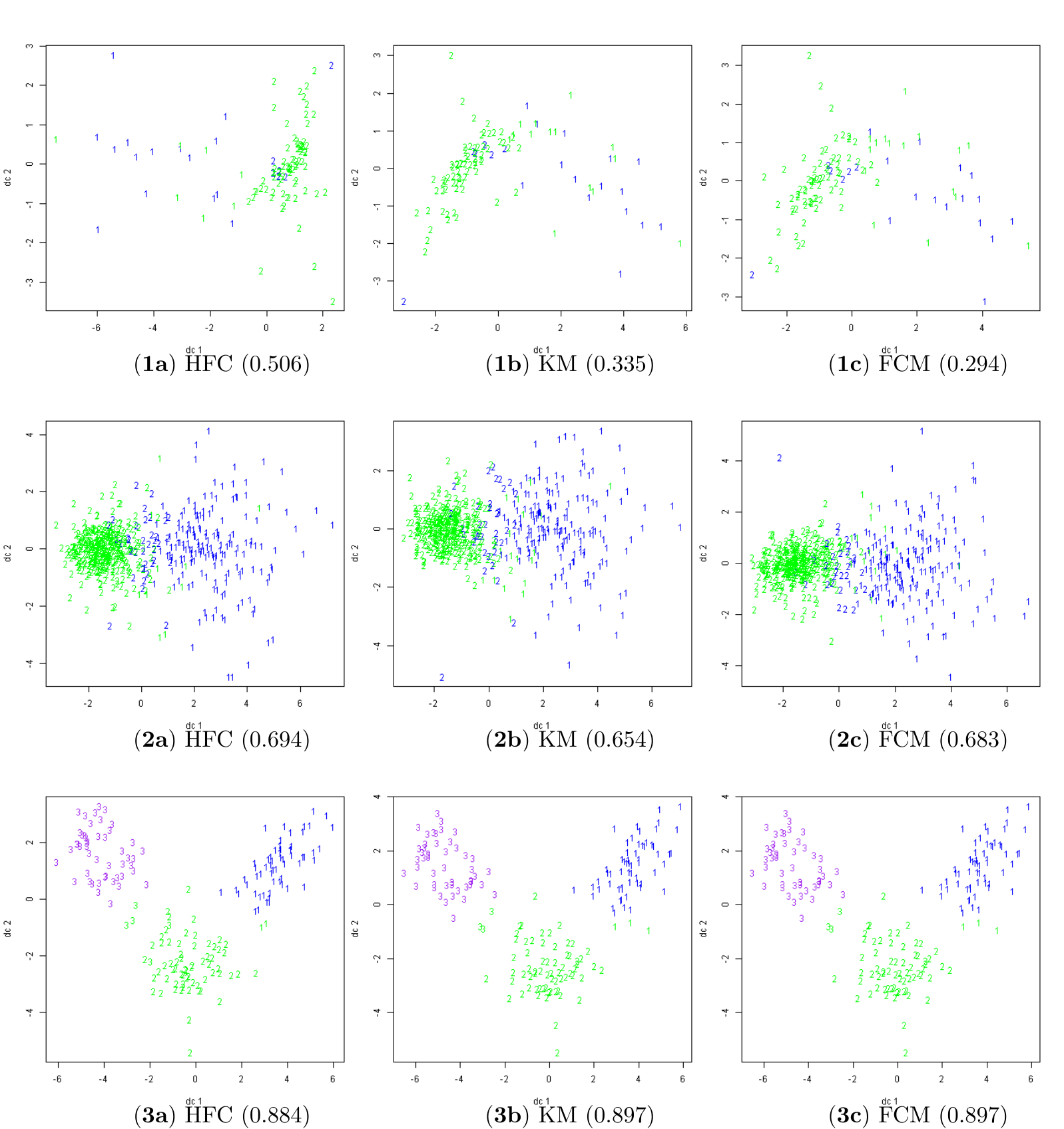} 
\caption{Cluster plots  of one run on imbalanced UCI datasets 1) Appendicitis with cluster distribution (0.2, 0.8) 2) WDBC with cluster distribution (0.37, 0.63) and 3) Wine with cluster distribution (0.33, 0.4, 0.27). The ARI values are reported in parentheses. \label{fig:app_seeds_wdbc}}
\end{figure}

The Appendicitis dataset is highly imbalanced and HFC outperformed the others in all four measures. Studying Figure \ref{fig:app_seeds_wdbc}  (1a) to (1c), we observed that many that belong to the green class are incorrectly assigned to cluster 1 for KM (1b) and FCM (1c).

For WDBC, there is considerable overlapping between the two clusters as shown in Figure \ref{fig:app_seeds_wdbc} (2a) to (2c). Despite this, HFC was able to overcome the overlapping while handling the imbalanced dataset, demonstrating HFC's ability to handle imbalanced datasets.

\begin{figure}[t]
\centering
  % \widefigure
% \hspace{-8pt}
\includegraphics[width=1.0\textwidth]{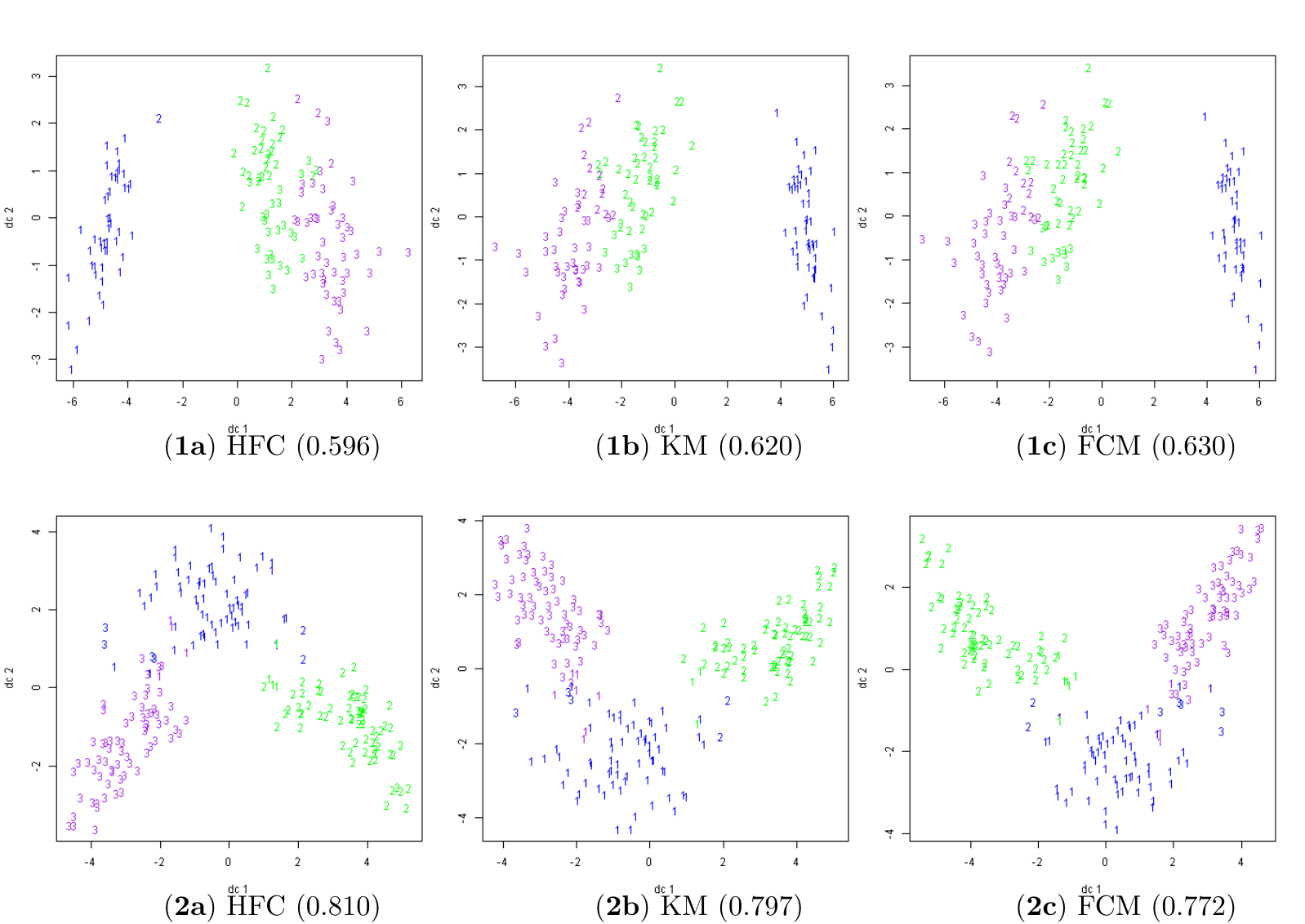}
\caption{Cluster plots  of one run on balanced UCI datasets 1) Iris and 2) Seeds both with cluster distribution (0.33, 0.33, 0.33). The ARI values are reported in parentheses. \label{fig:iris_seeds}}
\end{figure}

Despite Seeds being a balanced dataset, HFC outperformed KM and FCM. Studying the cluster plots on one run in Figure \ref{fig:iris_seeds} (2a) to (2c), those that belong to the purple group are more often incorrectly assigned to cluster 1 for KM and FCM than HFC.

From Table \ref{tab:UCIResults}, HFC did not outperform KM and FCM in Iris and Wine datasets. While ARI values showed competitive performance, with the near balanced and overlapping nature of the clusters, KM and FCM performed better on such dataset, particularly for Iris (see Figure \ref{fig:iris_seeds} (1a) to (1c)) where there is more overlapping between the green and purple groups.

\begin{landscape} 
\begin{table*}
\renewcommand{\arraystretch}{1.2}
\centering
\caption{Evaluation of algorithms on UCI datasets. Their cluster distribution are indicated in parentheses. \label{tab:UCIResults}}
\setlength{\tabcolsep}{3pt} 
{\begin{tabular}{l c c c c c c c c c c c}
\hline
	&	& \multicolumn{3}{c}{ARI}			& \multicolumn{3}{c}{Entropy}			&  \multicolumn{3}{c}{Dunn2}\\		
Dataset	&	&HFC	&KM	&FCM	&HFC	&KM	&FCM	&HFC	&KM	&FCM	\\
\noalign{\smallskip}
\hline
Appendicitis	&Mean	&\textbf{0.506}	&0.3	&0.294	
                        &\textbf{0.511}	&0.607	&0.604	%entropy
                        &\textbf{1.731}	&1.47	&1.475\\
	
(0.2, 0.8)	&StDev	&0	&0.05	&0	
	        &0	&0.029	&0	
	        &0	&0.054	&0\\
\\[0.5pt]

Iris	&Mean	&0.608	&0.62	&\textbf{0.63}	
                &\textbf{1.046}	&1.097	&1.098	%entropy
                &1.478	&\textbf{1.6}	&1.578\\

(0.33, 0.33, 0.33)
            &StDev	&0.016	&0	&0	
	        &0.018	&0	&0	
	        &0.06	&0	&0\\
\\[0.5pt]

Seeds	&Mean	&\textbf{0.793}	&0.784	&0.772	
                &1.098	&1.098	&1.098 %entropy	
                &1.611	&1.635	&\textbf{1.641}\\

(0.33, 0.33, 0.33)	&StDev	&0.024	&0.012	&0	
	        &0.001	&0	    &0	
	        &0.048	&0.019	&0\\
\\[0.5pt]

WDBC	&Mean	&\textbf{0.694}	&0.663	&0.683	
                &\textbf{0.631}	&0.638	&0.647 %entropy	
                &\textbf{1.212}	&1.196	&1.184\\

(0.37, 0.63)	&StDev	&0	&0.009	&0	
	        &0	&0.003	&0	
	        &0	&0.007	&0\\
\\[0.5pt]

Wine	&Mean	&0.894	&0.897	&0.897	
                &1.095	&1.093	&1.093	%entropy
                &1.255	&1.287	&1.287\\

(0.33, 0.4, 0.27)	&StDev	&0.023	&0	&0	
	        &0.001	&0	&0	
	        &0.004	&0	&0\\
\hline 
\end{tabular}}
\end{table*}
\end{landscape}

\subsection{Convergence of the HFC algorithm}
\begin{figure}[t]
\centering
  % \widefigure
% \hspace{-8pt}
\includegraphics[width=1.0\textwidth]{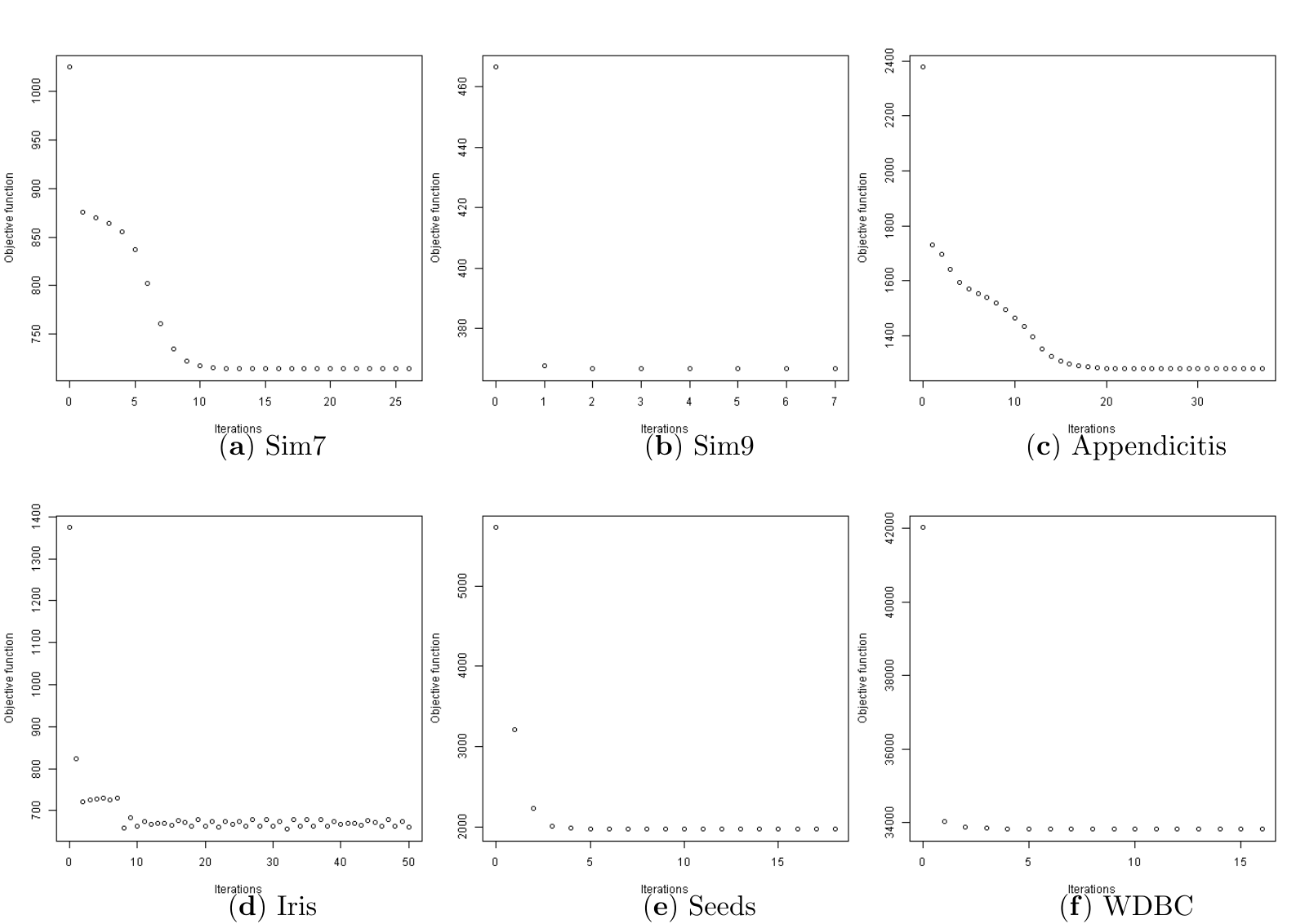}
\caption{Convergence plot of one run on selected datasets. \label{fig:convergence}}
\end{figure}

In Figure \ref{fig:convergence}, we presented the fitness value calculated by the objective function against the iterations for selected datasets which HFC perform both well and poorly in. The datasets are Sim7, Sim9, Appendicitis, Iris, Seeds and WDBC. We observed that HFC convergences at certain iterations except for HFC on Iris which has reached the maximum iteration of 50.

\section{Conclusion}
%\begin{itemize}
%\item Introduced HFC.
We introduced a hybrid fuzzy-crisp (HFC) clustering approach to solve the problem of centers of other small clusters being pulled towards outstandingly larger clusters. 
%\item Gave the algorithm and geometric interpretation.
In the HFC algorithm, the membership of a data point to a cluster becomes exactly zero when the data point is deemed ``far'' from the center. We showed that the HFC clustering can be interpreted as the problem of findind the closest point in a $(c-1)$-simplex from the origin in the $c$-dimensional Euclidean space, and proved the correctness of the algorithm.
%\item Tested with simulated and real data. (Did we confirm the original intent? That is, fuzzy clustering that works with highly biased dataset.)
The HFC algorithm was then tested on simulated and real data of varying cluster sizes and degrees of overlapping between clusters. Based on the results and analysis, HFC has been demonstrated to perform better on imbalanced datasets and can be competitive with KM and FCM on more balanced datasets. This was particularly evident on the two-cluster dataset with the mix ratios of (0.9, 0.1) and (0.8,0.2) where HFC outperformed KM and FCM. On dataset with mix ratios of (0.7,0.3), HFC can outperform KM and FCM where degrees of overlapping is small (average overlap $\leq 0.01$). Through evaluation measures such as ARI and Entropy as well as visual inspection via t-SNE biplots, we observed such performance from HFC.

%\item What's more to be done?
The experiments revealed the relative weakness of HFC in handling datasets that are well-balanced and/or largely overlapping, compared to other methods. One possible way to overcome this weakness may be to optimize the relative weights between the linear and quadratic terms ($\beta$ and $\{\alpha_i\}$, respectively) in the objective function (Eq. \ref{eq:JFCC}). This is left for future studies. 
%Further empirical analysis using more real-world imbalance datasets is needed for an exhaustive evaluation of HFC. Also, more detailed sensitivity analysis can provide deeper insights to HFC.

%\end{itemize}

\section*{Acknowledgments}
A. R. K. thanks Nobuhiro Go and Yusuke Umatani for suggesting this problem. 
Software used in this study is available at https://github.com/arkinjo/fccm.

\section*{CRediT author statement}
AR Kinjo: Conceptualization, Methodology, Formal Analysis, Validation, Writing - Original draft preparation, Writing - Reviewing and Editing, Visualization, Supervision

DTC Lai: Software, Validation, Investigation,  Data Curation, Writing- Original draft preparation, Writing - Reviewing and Editing

%\bibliography{../refs,../mypaper,../fcbiblo}
%\bibliographystyle{abbrvnat}

\end{document}